\def\paperID{14313}
\def\confName{ICCV}
\def\confYear{2025}

\def\paperTitle{TerraMind: Large-Scale Generative Multimodality for Earth Observation}

\def\authorBlock{
    Johannes Jakubik$^{1, \ast}$ \qquad
    Felix Yang$^{1,2 \ast}$ \qquad
    Benedikt Blumenstiel$^{1, \ast}$ \qquad
    Erik Scheurer$^{3}$ \qquad \\
    Rocco Sedona$^{3}$ \qquad 
    Stefano Maurogiovanni$^{3, 6}$ \qquad
    Jente Bosmans$^{4}$ \qquad
    Nikolaos Dionelis$^{4}$ \qquad \\
    Valerio Marsocci$^{4}$ \qquad 
    Niklas Kopp$^{1}$ \qquad
    Rahul Ramachandran$^{5}$ \qquad 
    Paolo Fraccaro$^{1, \dagger}$ \qquad \\
    Thomas Brunschwiler$^{1, \dagger}$ \qquad
    Gabriele Cavallaro$^{3, 6, \dagger}$ \qquad 
    Juan Bernabe-Moreno$^{1, \dagger}$ 
    Nicolas Longépé$^{4, \dagger}$ \qquad \\[5pt]
    $^{1}$IBM Research -- Europe \quad $^{2}$ETH Zurich\quad $^{3}$Forschungszentrum Jülich\\ $^{4}$European Space Agency $\Phi$-Lab\quad $^{5}$NASA IMPACT\quad $^{6}$University of Iceland \\
    {\tt\small johannes.jakubik1@ibm.com}\\[5pt]
}

\newif\ifreview 
\newif\ifarxiv 
\newif\ifcamera \newcommand{\cameraready}{\cameratrue}
\newif\ifrebuttal 

\cameraready 

\pdfoutput=1
\documentclass[10pt,twocolumn,letterpaper]{article}
\ifreview \usepackage[review]{iccv} \fi
\ifarxiv \usepackage[pagenumbers]{iccv} \fi
\ifrebuttal \usepackage[rebuttal]{iccv} \fi
\ifcamera \usepackage{iccv} \fi

\usepackage{graphicx}	
\usepackage{amsmath}	
\usepackage{amssymb}	
\usepackage{booktabs}
\usepackage{times}
\usepackage{microtype}
\usepackage{epsfig}
\usepackage{caption}
\usepackage{float}
\usepackage{placeins}
\usepackage{color, colortbl}
\usepackage{stfloats}
\usepackage{enumitem}
\usepackage{tabularx}
\usepackage{xstring}
\usepackage{multirow}
\usepackage{xspace}
\usepackage{url}
\usepackage{subcaption}
\usepackage{xcolor}
\usepackage[hang,flushmargin]{footmisc}
\usepackage{comment}
\usepackage{adjustbox}
\usepackage{pifont}

\ifcamera \usepackage[accsupp]{axessibility} \fi

% ...

\ifarxiv  \fi

\newcommand{\R}[1]{{%
    \textbf{%
        \ifstrequal{#1}{1}{\textcolor{red}{R#1}}{%
        \ifstrequal{#1}{2}{\textcolor{blue}{R#1}}{%
        \ifstrequal{#1}{3}{\textcolor{magenta}{R#1}}{%
        \ifstrequal{#1}{4}{\textcolor{teal}{R#1}}{%
                           \textcolor{cyan}{R#1}%
        }}}}%
    }%
}}

\newcolumntype{C}{>{\centering\arraybackslash}X}

\usepackage{xr-hyper}

\makeatletter
\newcommand*{\addFileDependency}[1]{
  \typeout{(#1)}
  \@addtofilelist{#1}
  \IfFileExists{#1}{}{\typeout{No file #1.}}
}

\makeatother
\newcommand*{\myexternaldocument}[1]{
    \externaldocument{#1}
    \addFileDependency{#1.tex}
    \addFileDependency{#1.aux}
}

\definecolor{iccvblue}{rgb}{0.21,0.49,0.74}
\usepackage[pagebackref,breaklinks,colorlinks,allcolors=iccvblue]{hyperref}
\usepackage[capitalize]{cleveref}
\crefname{section}{Sec.}{Secs.}
\crefname{table}{Table}{Tables}
\crefname{figure}{Fig.}{Figs.}

\ifarxiv \crefname{appendix}{App.}{Apps.}
\else \crefname{appendix}{Suppl.}{Suppls.} \fi

\unless\ifarxiv \myexternaldocument{_supplementary} \fi

\begin{document}
\title{\paperTitle}
\author{\authorBlock}

\twocolumn[{
\maketitle
\begin{center}
    \centering
    \includegraphics[width=1.0\linewidth]{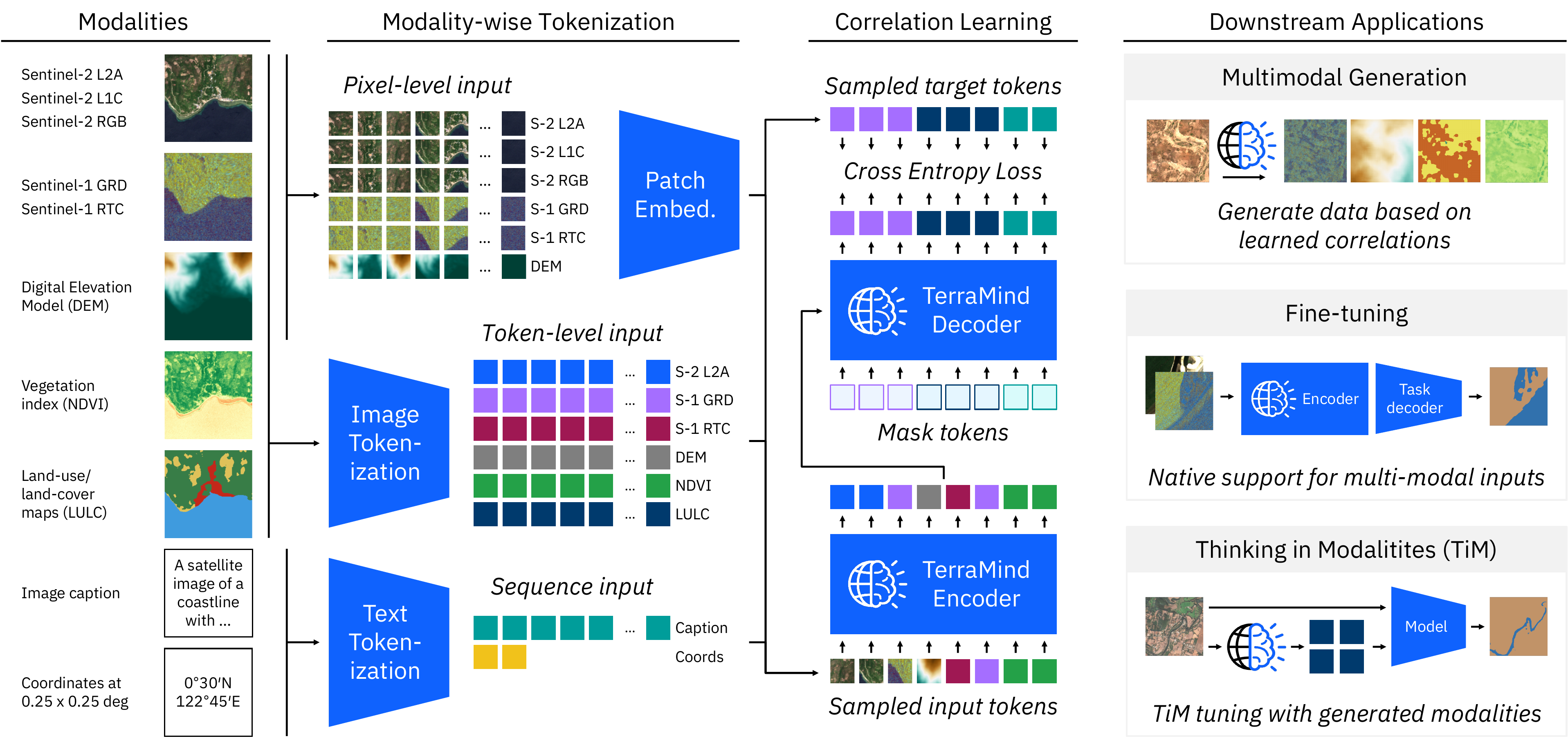}
    \captionof{figure}{TerraMind represents the first any-to-any generative, and large-scale multimodal model for Earth observation pre-trained on 500 billion tokens from global geospatial data. The model digests multi-scale representations at pixel-level and token-level simultaneously. TerraMindv1 unlocks (i)~generation, (ii)~zero-shot and finetuning applications, and (iii)~"Thinking-in-Modalities" finetuning and inference.}
    \label{fig:motivational_example}
\end{center}
}]
\ifcamera \renewcommand{\thefootnote}{}\footnotetext{$\ast$ Equal contribution} \fi
\ifcamera \footnotetext{$\dagger$ Equal supervision} \fi

\begin{abstract}
We present TerraMind, the first any-to-any generative, multimodal deep learning model for Earth observation (EO). Unlike other approaches, TerraMind is pretrained on dual-scale representations combining both token-level and pixel-level data across modalities. On a token level, TerraMind encodes high-level contextual information to learn cross-modal relationships, while on a pixel level, TerraMind leverages fine-grained representations to capture critical spatial nuances. In this paper, we demonstrate that (i)~TerraMind achieves beyond state-of-the-art performance in community-standard benchmarks, (ii)~TerraMind can leverage ``thinking in modalities'' (TiM)---the capability of generating additional artificial data during finetuning and inference to improve the model output---and (iii)~TerraMind's dual-scale early fusion approach results in well-structured embedding spaces. \color{black}Models and code have been open-sourced at \small\url{https://huggingface.co/ibm-esa-geospatial}\normalsize  and \small\url{https://github.com/ibm/terramind}\normalsize.\normalcolor
\end{abstract}

\section{Introduction}
\label{sec:intro}

Earth observation (EO) increasingly benefits from multimodality because of the important integration of complementary information from different data sources. This becomes particularly relevant as EO is spatiotemporally sparse due to low revisiting times or weather phenomena like cloud coverage. Vice versa, for computer vision, EO data is an important playground for the development of new approaches as there is significant publicly available data of very high quality and complexity. The available modalities range from sensors of different satellite missions to relevant complementary information like digital elevation. 

In this work, we introduce {TerraMind} as the first any-to-any generative multimodal model for EO. With TerraMind, we introduce a dual-scale pretraining on pixel-level and token-level and demonstrate benefits over training primarily on tokens. TerraMind encodes high-level contextual information in tokens to enable correlation learning and scaling, while, additionally capturing important fine-grained representations using pixel-level inputs. During pretraining, TerraMind predicts masked target tokens so that our pretraining objective boils down to a cross-modal patch classification problem that results in high-quality latent spaces.
TerraMind is pretrained on a custom global-scale geospatial dataset named TerraMesh with nine million samples that have been aligned spatiotemporally and across modalities \cite{terramesh}. In addition to radar and optical satellite images of the Copernicus Sentinel-1 (S-1) and Sentinel-2 (S-2) missions, our dataset contains task-specific modalities such as land use/land cover (LULC) and normalized difference vegetation index (NDVI) maps, metadata like digital elevation models (DEM) and geographic coordinates, and natural language in the form of captions. To the best of our knowledge, TerraMind represents the first truly generative, multimodal deep learning model for EO. 
\color{black}Additionally, in contrast to other recent models that utilize masked autoencoders like \cite{nedungadi2024mmearth}, contrastive learning, or diffusion techniques, TerraMind uniquely demonstrates benefits of leveraging token-based pretraining for EO. \normalcolor

\begin{figure}[b]
    \centering
    \includegraphics[width=\linewidth]{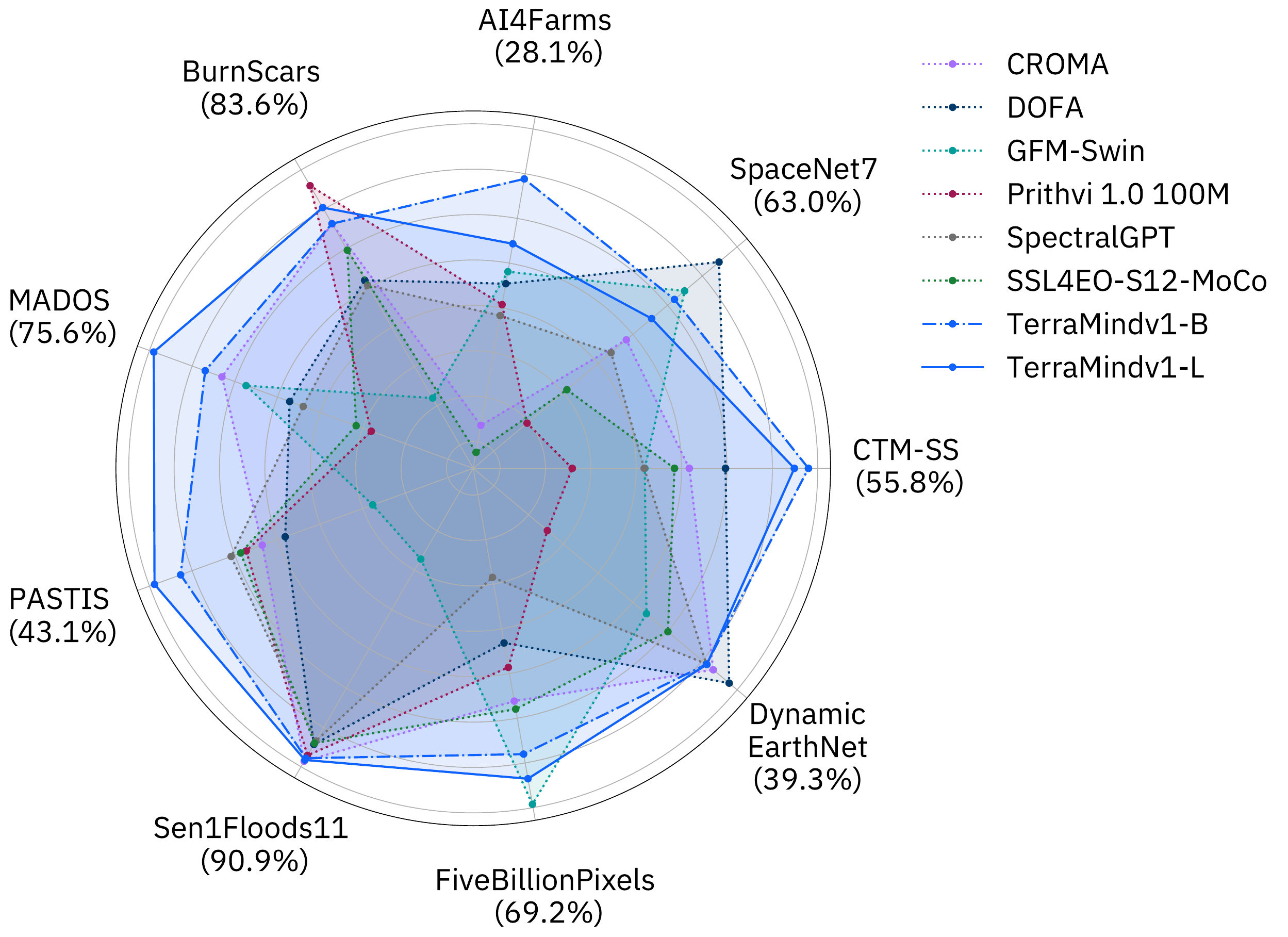}
    \caption{TerraMind outperforms other geospatial foundation models on PANGAEA benchmark \cite{marsocci2024pangaea} in finetuning. 
    Performance is measured in mIoU and min-max scaled per dataset.}
    \label{fig:radar}
\end{figure}

\begin{figure*}[h]  
    \centering 
    \includegraphics[width=1\linewidth]{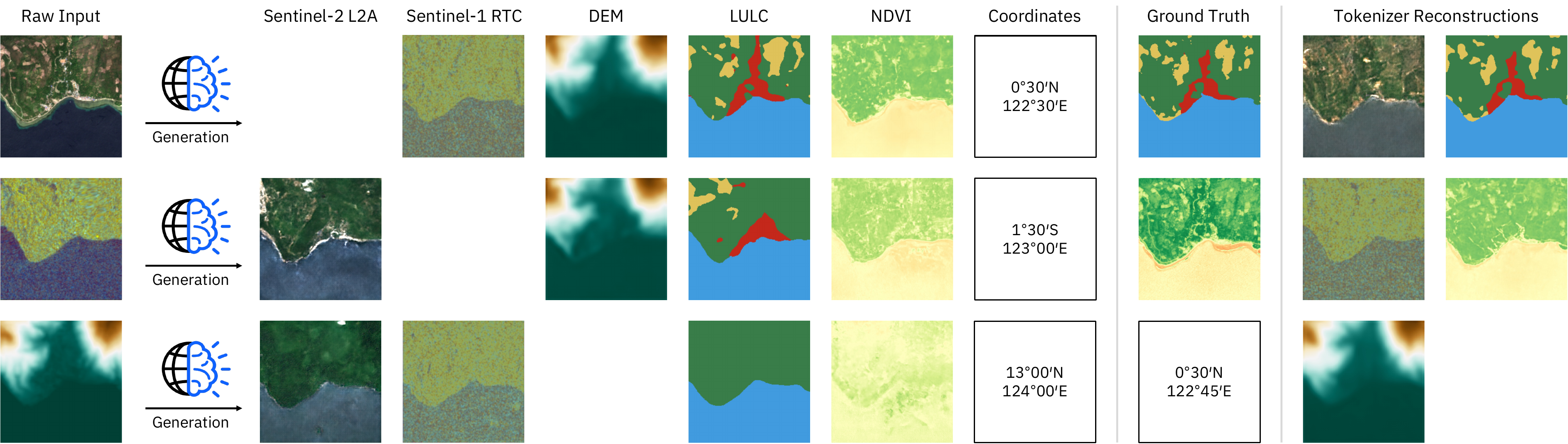}
    \caption{Chained generation example of TerraMindv1-B starting from either optical, radar, or digital elevation data. Left is input, middle is artificially generated data by TerraMind, right represents ground truths and tokenizer reconstructions, respectively.}
    \label{fig:generation-example}
\end{figure*}

We provide an overview of TerraMind's performance in a community-standard benchmark \cite{marsocci2024pangaea} in Figure~\ref{fig:radar} and highlight the any-to-any generative capabilities of TerraMind in Figure~\ref{fig:generation-example}.
Our key contributions are as follows: (i)~We introduce a dual-scale approach for generative multimodal pre-training leveraging data on pixel-level and token-level, which outperforms other fusion approaches and enhances embedding space structures. (ii)~We introduce \textit{thinking in modalities} -- similar to chain-of-thought approaches in LLMs -- for multi-modal models in EO, demonstrating that infusing generated data during finetuning improves the performance. (iii)~We demonstrate that TerraMind outperforms other geospatial foundation models both in unimodal and multimodal settings.

\section{Related Work}

\textbf{Computer vision in Earth observation.} Computer vision (CV) has significantly advanced EO \cite{zhu2017deep}. Many CV techniques, originally developed for natural image processing, have been adapted to EO \cite{tuia2023artificial}, often with minimal modifications. A wide range of tasks benefit from these methods, including classification \cite{dimitrovski2023current}, semantic segmentation \cite{yuan2021review} (e.g., land cover mapping \cite{vanetten2019spacenet, fibaek2024phileo}), change detection \cite{shafique2022deep} (e.g., disaster response \cite{durnov_xview2_2020}), object detection \cite{li2020object} (e.g., vessel identification \cite{paolo2022xview3sardetectingdarkfishing}), and regression (e.g., biomass estimation \cite{nascetti2023biomassters}). Deep learning architectures like CNNs \cite{zhao2022cnn} and Vision Transformers (ViTs) \cite{dosovitskiy2021imageworth16x16words} have demonstrated strong performance, often surpassing traditional remote sensing (RS) methods. However, EO presents unique challenges, including diverse sensor modalities \cite{astruc2024omnisat} and geospatial heterogeneity \cite{manas2021seasonal}. An emerging paradigm in EO is self-supervised learning (SSL) \cite{wang2022self} and geospatial foundation models (GFMs) \cite{visionpaper}, which aim to leverage vast amounts of unlabeled RS data to develop general purpose task models \cite{jakubik2023foundation}. While off-the-shelf CV models have shown promising results \cite{lahrichi2025self}, they do not fully exploit the unique characteristics of geospatial data. Many GFMs still rely on generic CV architectures \cite{mendieta2023gfm}, which were not explicitly designed to handle the complexities of EO, such as heterogeneous sensor sources (e.g., optical, radar, DEM) \cite{han2024bridging}, integrated with auxiliary data (e.g., text) \cite{liu2024remoteclip,marimo2025beyond}, and expert knowledge (e.g., prioritizing specific bands or indexes). In this direction, TerraMind better integrates domain-specific properties, developing a customized and expandable multimodal learning strategy.

\textbf{Multimodality in CV.} Multimodal CV is driven by the integration of diverse data streams \cite{yang2024mma}, such as natural images \cite{zhang2024vision}, natural language text \cite{bordes2024introduction}, temporal video data \cite{ren2024timechat}, and weather \cite{yang2024multi}, within large foundation models \cite{bommasani2021opportunities}. Starting from the alignment of images and texts \cite{clip}, these models moved beyond simple feature extraction, towards nuanced contextual understanding. The ability to combine several modalities allows for unprecedented capabilities in complex tasks \cite{han2024multimodal}, evidenced by the rapid advancement of multimodal Large Language Models (MLLMs) \cite{han2024multimodal}, that excel in tasks such as scene understanding \cite{cao2024maplm}, visual question answering \cite{driess2023palm}, and video analysis \cite{fu2024video}. Recent advances in architectures \cite{jain2024vcoder} and large scale pre-training \cite{cao2020behind} have enabled the development of models that learn highly effective cross-modal representations \cite{lin2023multimodality}, which can then be adapted to a wide variety of downstream tasks \cite{wu2025visionllm}.

\textbf{Multimodality in EO.} Multimodality in EO originates from data fusion and is typically understood as the integration of SAR and optical data \cite{chen2021self, hafner2022unsupervised, li2022deep, fuller2023croma} or the combination of optical data with vector data \cite{audebert_joint_2017}. Some studies have explored alternative combinations of data. In \cite{deuser2023sample4geo}, the authors introduce a contrastive framework for comparing RS images and street views. Even different optical sensors can be considered different modalities \cite{swope2021representation, marsocci2024crosssensor}. Similarly, several multi-view images (i.e. multimodal) datasets \cite{machado2020airound, garioud2023flair, nedungadi2024mmearth} are introduced. More recent approaches combined text and images \cite{li2024visionlanguage}, both for discriminative \cite{liu2024remoteclip} and generative \cite{khanna2023diffusionsat} purposes. Lately, different GFMs are trained in a multimodal way \cite{astruc2024omnisat, nedungadi2024mmearth, xiong2024neural}, still focusing either on a specific set of modalities (e.g., vision \cite{xiong2024neural}, \cite{astruc2024anysat}) or tasks (e.g., generative \cite{khanna2023diffusionsat}). \textcolor{black}{Compared to multi-scale high-quality generation models for optical data, like MetaEarth \cite{yu2024metaearth}, our approach allows to generate any modality from any other pretraining modality.} To the best of our knowledge, no existing model has combined a wide and diverse amount of modalities both for discriminative and generative purposes, as TerraMind does. We provide a comparison in Table~\ref{tab:arch_comparison}.

\begin{table}[h]       
    \centering    
    \small
    \setlength\tabcolsep{2pt} 
    \renewcommand{\arraystretch}{1.2}
    \begin{tabularx}{\linewidth}{l>{\hsize=1.5\hsize}C>{\hsize=.75\hsize}C>{\hsize=.75\hsize}C}
        \toprule
         Model & Modalities & Any-to-Any Generation & Multi-Scale Features \\
         \midrule
         RemoteCLIP & optical, text & \ding{55} & \ding{55}\\
         CROMA & optical, radar & \ding{55} & \ding{55}\\
         AnySat & aerial, optical, radar, NAIP & \ding{55} & \ding{55}\\
         DeCUR & optical, radar & \ding{55} & \ding{55}\\
         DOFA & optical, radar, hyperspectral, NAIP & \ding{55} & \ding{55}\\
         \textcolor{black}{MetaEarth} & \textcolor{black}{optical (unimodal)} & \textcolor{black}{\ding{55}} & \textcolor{black}{\ding{51}}\\
         Galileo & optical, radar, elevation, weather, location, population, ... & \ding{55} & \ding{51}\\
         \midrule
         \textbf{TerraMind} & \textbf{optical, radar, land use, elevation, vegetation index, location, text} & \ding{51} & \ding{51} \\

         \bottomrule
    \end{tabularx}
    \caption{Comparison of TerraMind to other model architectures. TerraMind represents a first-of-its-kind generative, multimodal model.}
    \label{tab:arch_comparison}
\end{table}

\label{sec:related}

\section{Dataset}
\label{sec:dataset}

For the pretraining of TerraMind and its tokenizers, we create a multimodal dataset called TerraMesh \cite{terramesh}, which will be open-sourced to the community. TerraMesh builds on existing datasets, which we expand by adding modalities from external data sources or by applying pseudo-labeling. We provide an overview of the aligned image modalities and a detailed dataset description in the supplementary material.

\textbf{Base datasets.} TerraMesh is based on SSL4EO-S12 \cite{wang2023ssl4eo,blumenstiel2025ssl4eos12v11} and MajorTOM-Core \cite{francis2024major}, two unlabeled remote sensing datasets containing co-aligned radar and optical imagery from Sentinel-1 and Sentinel-2 satellites. SSL4EO-S12 has lower geographic coverage but is multi-seasonal. MajorTOM-Core covers most of the Earth's land surface at a single timestamp. For MajorTOM-Core, we apply a subsampling scheme based on LULC classes and ecoregions. TerraMesh includes a total of approximately 9 million globally distributed samples from both Sentinel-1 and Sentinel-2, each measuring 264$\times$264 pixels at 10m resolution.

\textbf{Additional modalities.} We obtain co-aligned yearly LULC maps by ESRI\footnote{\tiny\url{https://planetarycomputer.microsoft.com/dataset/io-lulc-annual-v02}} with nine land use classes. Additionally, we leverage SEnSeI v2 \cite{francis2024sensor} as a cloud and ice annotation model and update the ESRI LULC classes for better spatiotemporal alignment. NDVI maps are computed using the corresponding spectral bands from Sentinel-2. DEM is extracted from the Copernicus DEM 30m dataset \cite{dem2022}, which provides global coverage of the Earth's elevation at a 30m resolution. 
Captions are generated synthetically by constructing RGB images from Sentinel-2 patches using the corresponding spectral bands and processing them with LLaVA-Next~\cite{li2024llavanext}. A tailored prompt guides the model to describe the content of each image as described in \cite{marimo2025beyond}.
For geolocations, we round latitude and longitude from the center of each patch to the nearest quarter degree and store the discretized coordinates as strings in a pre-defined format.


\section{Methods}
\label{sec:methods}

TerraMind pretraining is two-staged following \cite{mizrahi20234mmassivelymultimodalmasked}. We first pretrain unimodal tokenizer models, tokenize the modalities, and then leverage token-level and pixel-level input to pretrain the TerraMind encoder-decoder architecture. We describe those individual stages in the following.

\subsection{Tokenization}
We develop modality-specific tokenizers to encode each modality into a sequence of discrete tokens for pretraining and decode token sequences back to images. Thus, TerraMind is in principle compatible with any modality, as long as it can be tokenized and aligned with other modalities. For reasons of space, we delegate most experiments related to the tokenizer performances to the supplementary material.

\begin{figure}[htb]
    \centering
    \includegraphics[width=1\linewidth]{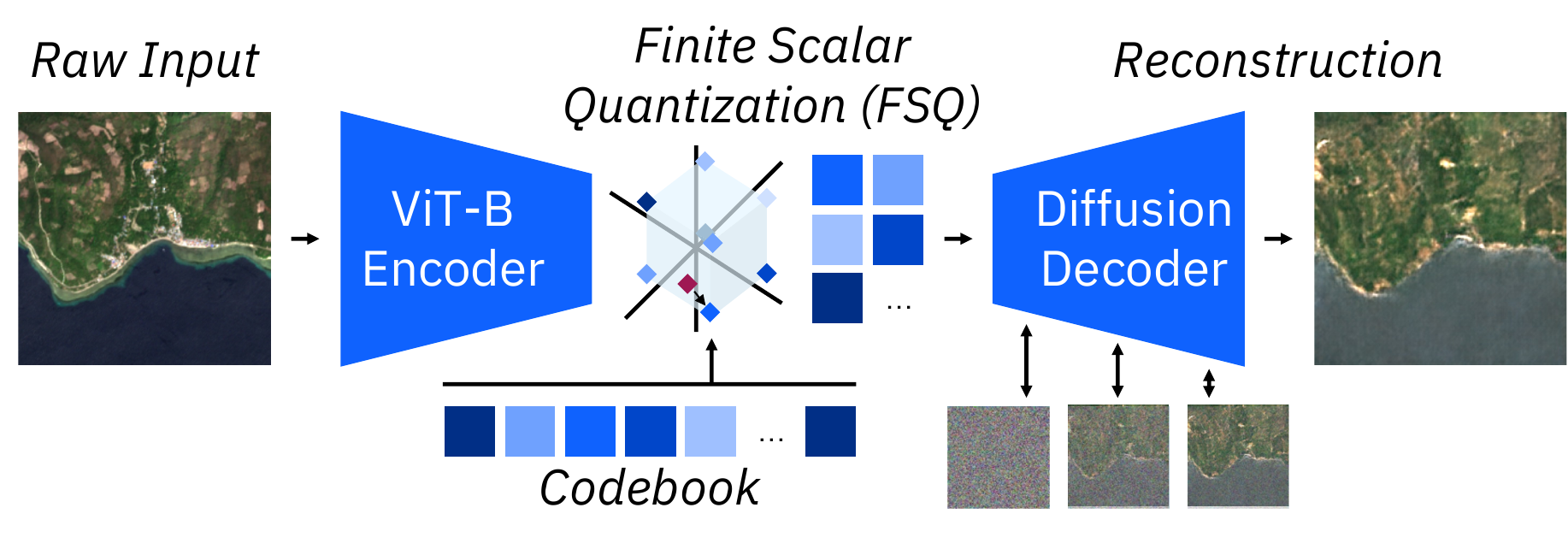}
    \caption{Tokenizer for image-like modalities combining finite-scalar quantization \cite{mentzer2023finite} with diffusion decoding.}
    \label{fig:tokenizer-arch}
\end{figure}

\textbf{Image-like modalities.} We train autoencoder-based architectures with a quantization step in the bottleneck for image-like modalities such as S-1, S-2, LULC, NDVI, and DEM. Tokenizer encoders process an input image and generate a latent representation for each 16$\times$16 patch, which is then discretized with finite-scalar-quantization (FSQ) \cite{mentzer2023finite} into one of $N$ codewords. All tokenizers use a vocabulary size of 16K besides the simpler LULC modality for which we use 4K. These codewords are then used by the diffusion decoder to reconstruct the original image. \textcolor{black}{The benefit of leveraging diffusion decoders lies in facilitating cross-modal generation in TerraMind by transforming tokens back into images.} By mapping each codeword to a unique integer in $\{0, 1, \dots, N-1\}$, we obtain discrete tokens for each image patch. We pretrain the tokenizers in a self-supervised setting. FSQ as quantization method enhances training stability  \cite{mentzer2023finite} compared to vector quantization \cite{van2017neural} by eliminating the need for codebook-related loss terms. Notably, FSQ is heavily influenced by ideas of neural compression \cite{gomes2025lossy}. For example, on 12-band S-2 images, we achieve compression rates of over 3000x by applying quantization. We summarize the architecture of our tokenizers in Figure~\ref{fig:tokenizer-arch}. \color{black}
The main objective of the overall tokenizer is to encode image patches consistently into discrete tokens based on semantic similarity to enable cross-modal correlation learning. Therefore, the loss of some details is an expected trade-off, since the focus is on grouping similar patches rather than preserving all fine-grained features. Naturally, more accurate reconstructions facilitate cross-modal generation, however the main focus of the pretraining lies on consistent cross-modal correlation learning. We provided further details on the pretraining of the tokenizers in the supplementary material.
\normalcolor

\textbf{Sequence-like modalities.} We treat both captions and geolocations as text and use a single text tokenizer to process both modalities. By discretizing the geographic coordinates and representing them as strings, we introduce special coordinate tokens into the vocabulary. This allows us to encode geolocations as a sequence of discrete tokens, beginning with a latitude token followed by a longitude token. For textual data, we modify the existing WordPiece tokenizer \cite{kenton2019bert}. 

\subsection{Pre-training}

\textbf{Architecture.} TerraMind uses a symmetric Transformer-based encoder-decoder architecture proposed in  \cite{mizrahi20234mmassivelymultimodalmasked}, which is designed to process sequences of multimodal tokens. In addition to discrete tokens, TerraMind accepts pixel-level inputs, specifically satellite imagery and digital elevation maps. For pixel-level inputs, we apply learnable patch-wise linear projections to generate patch embeddings for each 16$\times$16 patch, similar to the approach used in ViT \cite{dosovitskiy2021imageworth16x16words}. 

\begin{figure}[htb]                 
    \centering    
    \includegraphics[width=1\linewidth]{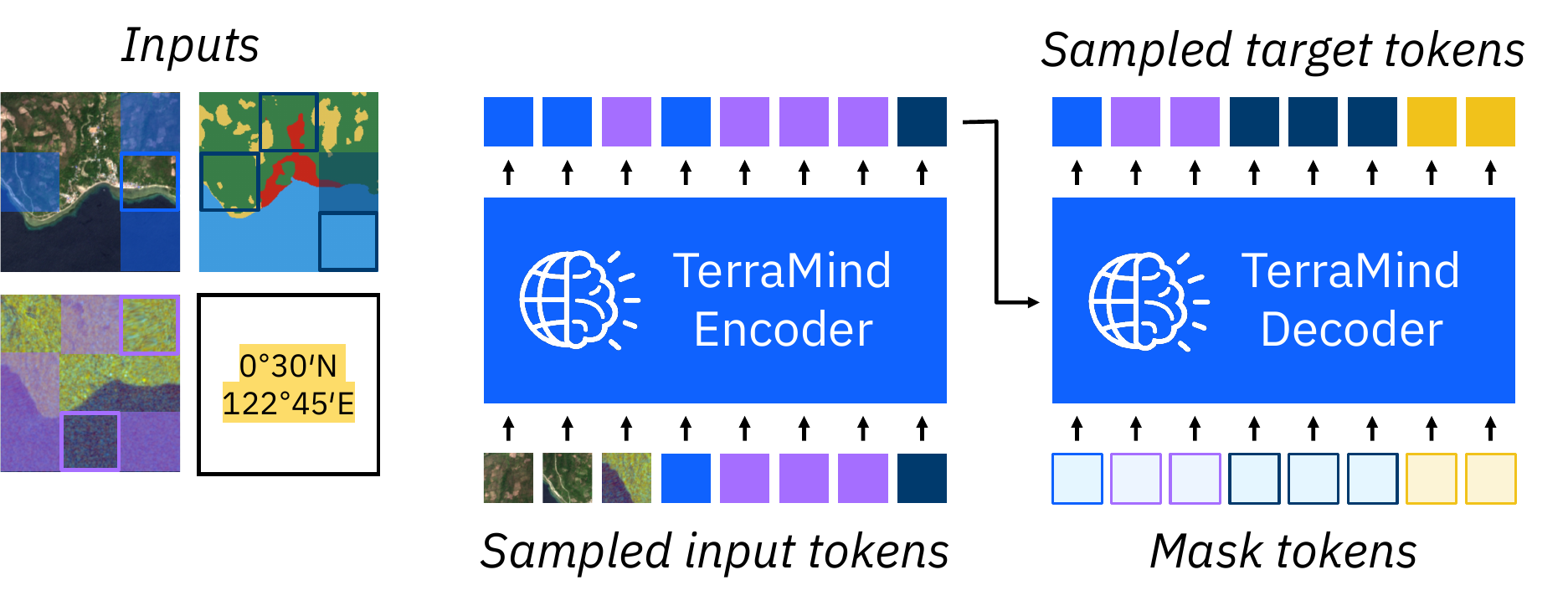}  
    \caption{Illustration of the pre-training task. Given an encoded multimodal sample of random subsets of patches and input tokens, the decoder predicts target tokens for the masked input.}  
    \label{fig:pre_training}  
\end{figure}

\color{black}
\textbf{Dual-scale early fusion.} 
In contrast to \cite{mizrahi20234mmassivelymultimodalmasked}, we not only embed token-level data but additionally leverage pixel-level data across a range of input modalities to introduce a dual-scale feature representation to support the structuring of the embedding space. 
Both tokens and patches represent a 16x16 pixel area. Tokens represent this area via a single discrete integer value, while the image patches describe the same area with the actual floating point sensor data. Thus, during pretraining, the model not only learns a correlation \textit{between} modalities (i.e., cross-modal learning) but also between different levels of abstraction \textit{within} the same modality. The low-level token information enables cross-modal correlation learning, while adding pixel level input accounts for spatial nuances. Based on dual-scale features the model further learns to better structure pixel-level data in the embedding space via the corresponding information from the discrete token.
We illustrate the pretraining paradigm in Figure \ref{fig:pre_training}. The model is agnostic to processing tokens or patches in the input space, while the target is generally token-level data. We use six pixel-level modalities and eight token-level modalities. 
\normalcolor

\textbf{Masking strategy.} TerraMind applies a masked modeling approach in the token space following \cite{mizrahi20234mmassivelymultimodalmasked}. The model leverages a set of randomly selected target tokens that have to be reconstructed from a randomly selected set of input tokens and pixel-level data. During pre-training, we sample input and target data from a Dirichlet distribution.

We opt for masked token reconstruction to familiarize the model with the absence of entire modalities, which is crucial for a high usability of a multimodal model in Earth observation. During pre-training, the model learns an internal representation of unseen modalities which is expected to benefit a range of downstream applications. In addition, sampling input and target tokens improves the computational efficiency of the pre-training, as each token is a compressed representation of a patch with compression factors of between 250x and 3000x depending on the modality. Finally, without tokenized representations of the image-like modalities, it is challenging to learn the correlation to sequence-like modalities. The overall training objective of TerraMind boils down to a cross-modal patch-level classification problem optimized via a cross entropy loss:
\begin{equation}
    \mathcal{L}_{\text{CE}} = - \sum_{i=1}^{N} y_i \log(p_i),
\end{equation}
where \( y_i \) is the one-hot encoded true class of token \( i \), \( p_i \) is the predicted probability for token \( i \), \( N \) is the total number of possible tokens. 
Interestingly, we can infer an upper bound loss for a random model where the cross entropy loss will collapse to the natural logarithm of the vocabulary size $\mathcal{L}_{\text{CE, random}} = - \sum_{i=1}^{N} y_i \log\left(\frac{1}{N}\right) = \log(N)$.

\textbf{Scaling.} We trained three versions of TerraMind scaling across model size, compute, and data. In addition, we pre-train different versions of TerraMind with respect to the number of dual-scale features. \textbf{TerraMindv1-B} is pre-trained on 500B tokens for 6 days on 32 NVIDIA A100 GPUs. The model uses dual-scale features from both token-level and pixel-level. During initial experiments, we observed significant improvements from scaling model size when switching from a tiny backbone to a small backbone to a base backbone. Therefore, we pre-trained \textbf{TerraMindv1-L} on a large backbone with 500B tokens on 32 NVIDIA A100GPUs trained for 10 days. 
Finally, to better understand the effect of scaling across the dual-scale feature representation, we pre-train \textbf{TerraMindv1-B-single} as a single-scale model on primarily token-level data with optical S-2 L2A data as only pixel-level input (compared to pixel-level S-2 L1C, S-2 RGB, S-1 GRD, S-1 RTC, and DEM in TerraMindv1-B and -L). TerraMindv1-B-single is pretrained on 500B tokens from over one million samples for 6 days on 32 NVIDIA A100 GPUs. 
We summarize the scaling behavior in model size, compute, and data in Figure~\ref{fig:scaling} of the supplementary material. We additionally provide final validation losses in Table~\ref{tab:val_losses} comparing v1-B and v1-L with the theoretical random loss.

\subsection{Generation}
Once pretrained, TerraMind can generate tokens for any modality, conditioned on any subset of input modalities. The generative capabilities unlock various zero-shot tasks, such as water body segmentation. \color{black}For the generation of image-like modalities, the decoder receives mask tokens for the modality to be generated and predicts the corresponding tokens based on the encoded input. For sequence-like modalities, the decoder generates the output autoregressively. After generating tokens from the target modality, the corresponding tokenizer decoder allows to map from token-space to image or text space. TerraMind further supports chained generation which ensures consistency across generated modalities. The chained generation represents a conditional probability distribution where the prior probability distribution is determined by the input modality, and all subsequent modalities are generated conditioned on the input modality and potentially other generated modalities.

%

\textcolor{black}{\subsection{Thinking-in-Modalities}}
\noindent
\color{black}
Thinking in Modalities (TiM) is a recursive fine-tuning and inference technique designed to enhance multimodal learning by leveraging the generative capabilities of the model itself. Given an input $x \in \mathcal{X}$ (e.g., an optical satellite image), the model first generates additional synthetic modalities $\tilde{x} = f_{\text{gen}}(x)$ on a token-level using a learned generative function $f_{\text{gen}}$. These generated tokens are then concatenated with the original input and jointly processed by the downstream model $f$ (e.g., TerraMind encoder with a segmentation head), yielding the final output $y = f(x, f_{\text{gen}}(x))$. This formulation allows the model to reason over both observed and inferred modalities, effectively enriching the input space. TiM can leverage multiple generated modalities which are then generated in a chained approach. For example, for $k$ modalities, the input is augmented with newly generated modalities: 
\begin{equation}
    \tilde{x}^{(k+1)} = \tilde{x}^{(k)} \cup f_{\text{gen}}(\tilde{x}^{(k)}), 
\end{equation}
and the final model output is described by:
\begin{equation}
    y = f(\tilde{x}^{(K)}).
\end{equation}
This recursive augmentation mimics a chain-of-thought process, enabling the model to iteratively refine its internal representation, particularly in scenarios with missing modalities.
\normalcolor

\section{Experiments}
\label{sec:experiments}

In this section, we describe the performance gains resulting from TerraMind and experiment with the unlocked capabilities of any-to-any generation and Thinking-in-Modalities.

\subsection{Foundational experiments}
\textbf{Multimodality vs. unimodality.} As a first motivational experiment, we outline the benefit of using multimodal data in Earth observation at the example of water body mapping. Specifically, we leverage the ViT-B encoders from the unimodal tokenizer models for S-1, S-2, and LULC, concatenate their embeddings, and train a segmentation head with four ConvNeXt \cite{liu2022convnet2020s} blocks as a late fusion approach. The results in Table~\ref{tab:sen1floods11_uni_vs_multi} (left) suggest that regardless of which modalities we combine, the combination of two modalities always outperforms each unimodal model. Combining all three modalities achieves the best overall performance.

\begin{table}[h]   
    \centering 
    \footnotesize
    \begin{tabular}{lcc}
        \toprule
        Input & Late fusion & Token-level fusion\\
        \midrule
        S-1 & 61.01 & {63.94 (2.93pp$\uparrow$)}  \\
        S-2 & 72.70 & {76.32 (3.62pp$\uparrow$)}  \\
        LULC & 71.77 & 70.96 (0.81pp$\downarrow$)  \\
        S-1 + S-2 & 73.83 & {76.74 (2.91pp$\uparrow$)}  \\
        S-1 + LULC  & 73.86 & 73.76 (0.10pp$\downarrow$)  \\
        S-2 + LULC & 75.65 & {77.04 (1.39pp$\uparrow$)}  \\\midrule
        S-1 + S-2 + LULC & \textbf{76.00} & \textbf{76.88 (0.88pp$\uparrow$)}  \\
        \bottomrule
    \end{tabular}
    \caption{Water body mapping on Sen1Floods11 \cite{Bonafilia_2020_CVPR_Workshops} measured in IoU on water class. Model sizes and architectures are comparable. \textbf{Left column:} Late fusion of tokenizers. The average improvement of full multimodality over the individual unimodal performance is 7.5pp IoU. \textbf{Right column:} Finetuning results of TerraMindv1-B-single as a mid fusion approach based on masked correlation learning. Gains over late fusion in percentage points in parentheses.}
    \label{tab:sen1floods11_uni_vs_multi}
\end{table}

\textbf{Token-level fusion vs. late fusion.} In Table~\ref{tab:sen1floods11_uni_vs_multi} (right), we investigate the effects of fusing the inputs on a token level through masked token reconstruction. We observe that token-level fusion outperforms late fusion. The performance gains are particularly high when LULC data is not available. This suggests that early fusion captures an internal representation of the multimodal state---especially pronounced for LULC---that benefits fine-tuning. With those findings in mind, we will explore the effects of using additional multi-modal pixel-level input in a dual-scale pretraining in Section~\ref{sec:finetuning}.

\subsection{Generation experiments}    

%
%
%
%

TerraMind supports any-to-any generation. In the following, we provide examples of the generation performance starting from: (i)~an information-rich modality, like optical S-2 L2A data, and (ii)~minimal information based on the geolocation. In Figure~\ref{fig:generation-example}, we observe that TerraMind performs strongly in generating image-like modalities like S-1, LULC, and DEM from optical S-2 L2A data. \color{black} We provide a quantitative overview on the quality of the generations on unseen validation data in Table~\ref{tab:generation_performance}. 
Overall, we observe an interesting asymmetry in the generative performance of TerraMind where (a) radar-to-optical generation achieves reasonable quality in terms of SSIM and PSNR -- indicating structural and visual fidelity with some perceptual degradation -- and (b) optical-to-radar generation yields higher PSNR values but lower SSIM, suggesting visually plausible outputs that lack strong structural alignment. The quality of generated DEM suggests to be structurally very strong, but noisy. The errors for DEM generations suggest that the level of altitude is difficult to infer for the model. We compare these scores with the reconstruction quality of the auto-encoding tokenizers in the supplementary material that can serve as upper bounds. Additionally, we provide experiments on the generation quality using token-level instead of pixel-level inputs. Finally, we demonstrate the quality of generations at kilometer scale in Figures~\ref{fig:singapore-large_tile} and \ref{fig:santiago-large_tile}.
\normalcolor

\begin{figure}[htbp]
    \centering

    \begin{subfigure}[b]{0.23\textwidth}
        \includegraphics[width=\textwidth]{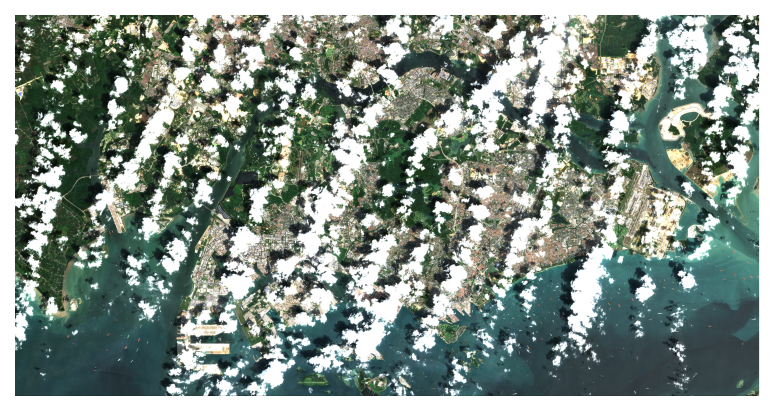}
        \caption{\textbf{Input}: S-2 L2A data capturing Singapore in January 2025.}
        \label{fig:rgb}
    \end{subfigure}
        \hfill
    \begin{subfigure}[b]{0.23\textwidth}
        \includegraphics[width=\textwidth]{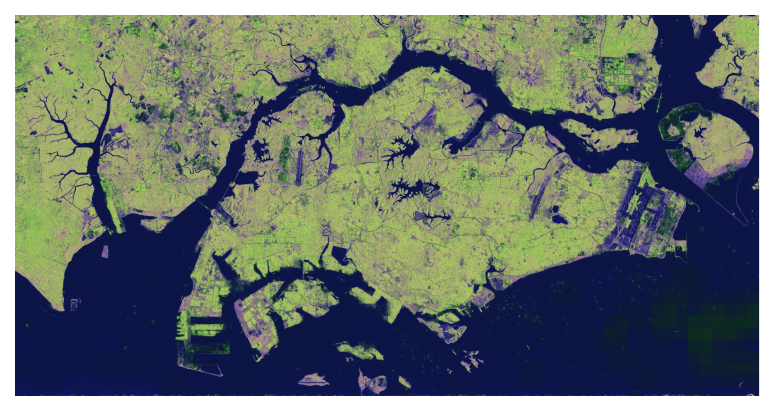}
        \caption{\textbf{Generation}: S-1 RTC composition generated by TerraMind.}
        \label{fig:grdvh}
    \end{subfigure}

    \vspace{1em}

    \begin{subfigure}[b]{0.23\textwidth}
        \includegraphics[width=\textwidth]{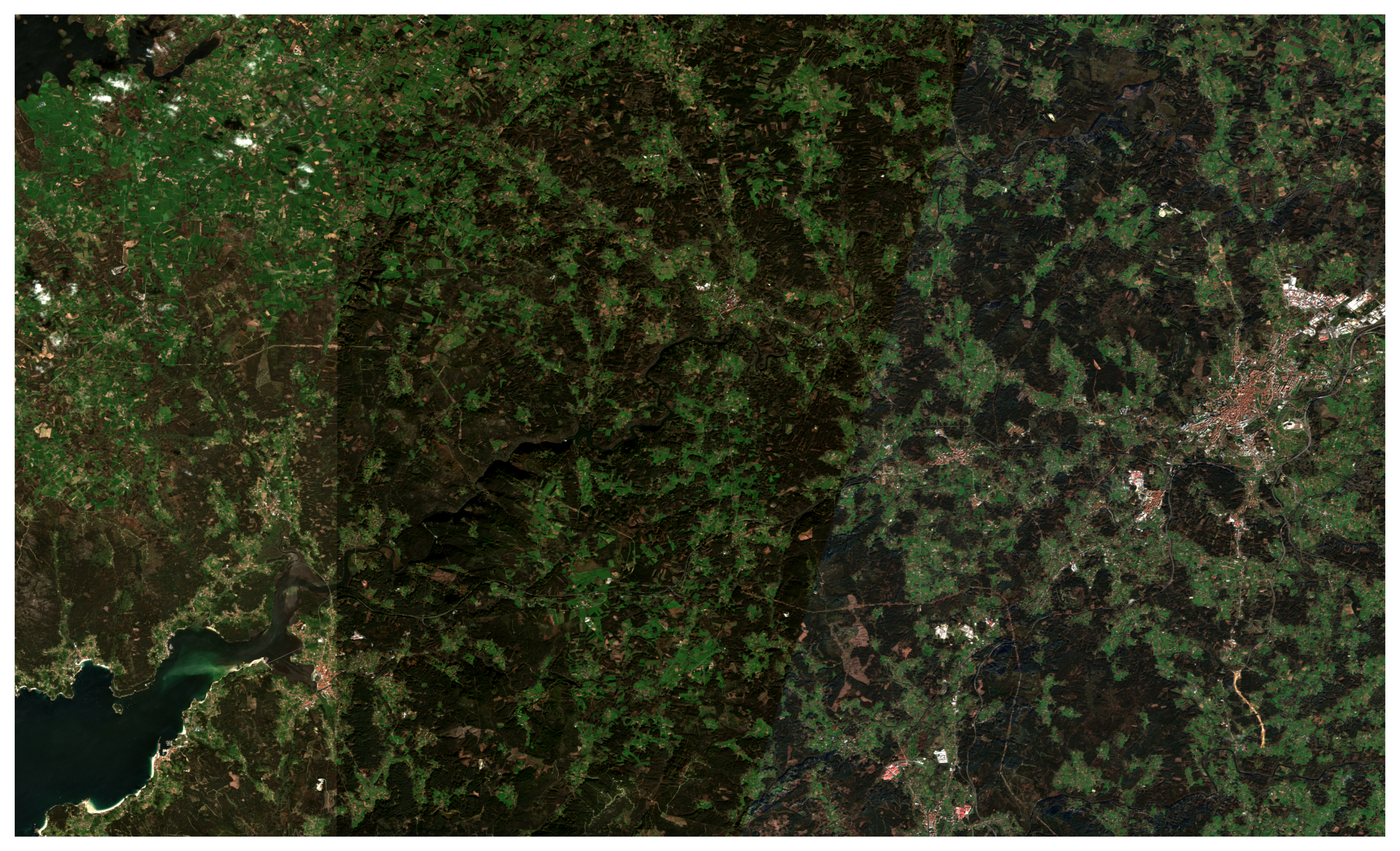}
        \caption{\textbf{Input}: S-2 L2A data capturing Northern Spain in January 2025.}
        \label{fig:grdvv}
    \end{subfigure}
    \hfill
    \begin{subfigure}[b]{0.23\textwidth}
        \includegraphics[width=\textwidth]{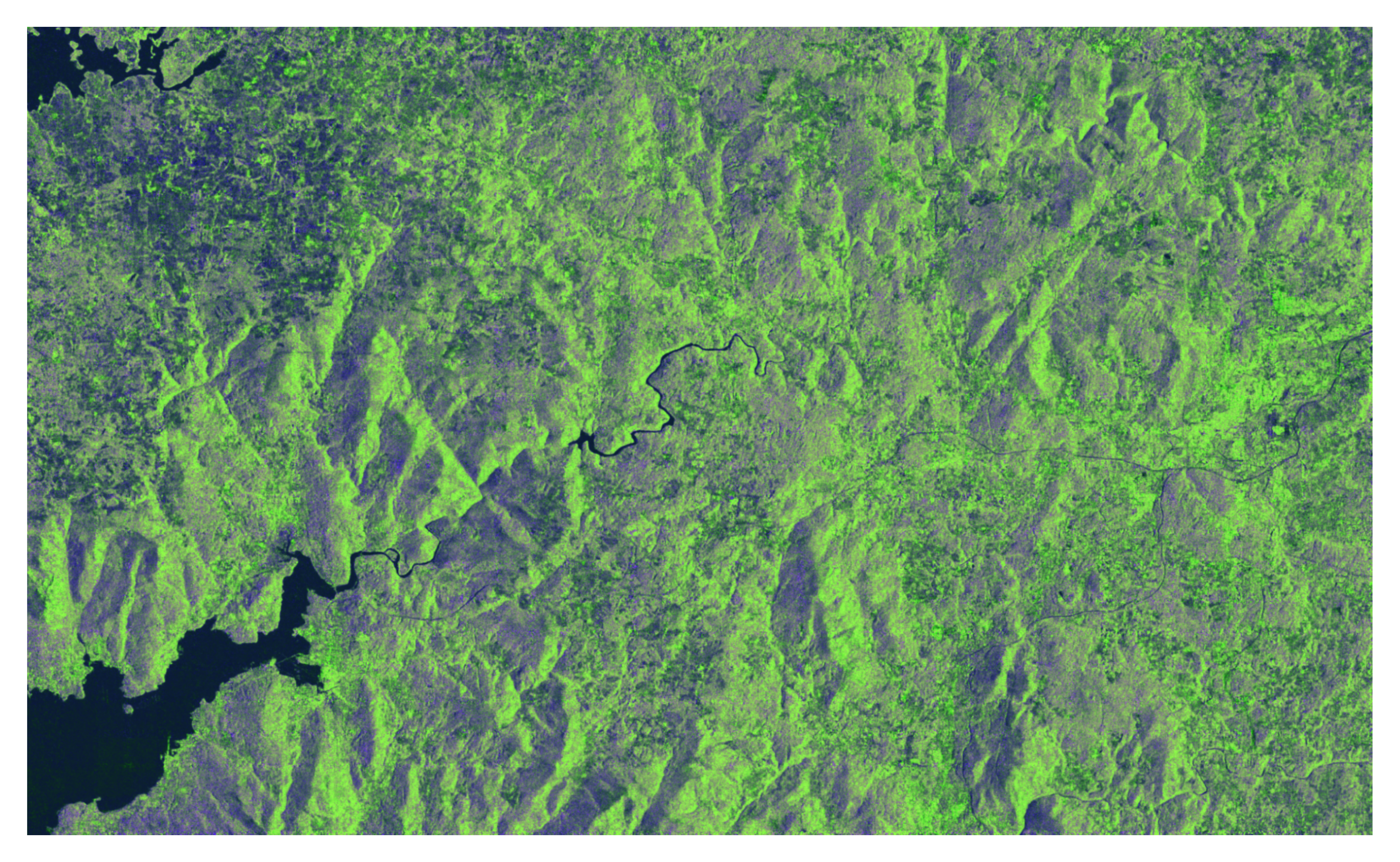}
        \caption{\textbf{Generation}: S-1 GRD composition generated by TerraMind.}
        \label{fig:rtcvh}
    \end{subfigure}

    \caption{Generated S-1 imagery using TerraMind. We provide large-scale visualizations in the supplementary material.}
    \label{fig:2x3_grid}
\end{figure}

\begin{table}[htb]
\centering
\footnotesize
\color{black}
\arrayrulecolor{black}
\begin{tabular}{lcccc}
\toprule
\textbf{Modalities} & \textbf{MAE  $\downarrow$} & \textbf{RMSE$\downarrow$} & \textbf{SSIM$\uparrow$} & \textbf{PSNR$\uparrow$} \\
\midrule
S-1 GRD $\rightarrow$ S-2 L2A & 0.074 & 0.116 & 0.750 & 26.210 \\
S-1 GRD $\rightarrow$ DEM & 163.0 & 320.8 & 0.878 & 20.694 \\
S-1 GRD $\rightarrow$ NDVI  & 0.180 & 0.225 & 0.438 & 18.990 \\
\midrule
S-1 RTC $\rightarrow$ S-2 L2A & 0.113 & 0.194 & 0.695 & 24.251 \\
S-1 RTC $\rightarrow$ DEM & 298.8 & 799.2 & 0.873 & 20.009 \\
S-1 RTC $\rightarrow$ NDVI  & 0.172 & 0.211 & 0.465 & 19.529 \\
\midrule
S-2 L2A $\rightarrow$ S-1 GRD & 2.942 & 3.877 & 0.531 & 28.678 \\
S-2 L2A $\rightarrow$ S-1 RTC & 2.636 & 3.391 & 0.430 & 28.993 \\
S-2 L2A $\rightarrow$ DEM & 215.8 & 745.5 & 0.942 & 20.616 \\
\bottomrule
\end{tabular}
\caption{\color{black}Quantitative evaluation of generations on unseen global validation dataset using 10 diffusion steps. MAE and RMSE metrics are in physical units: meter (DEM), reflectance (S-2), and db (S-1). \normalcolor}
\label{tab:generation_performance}
\end{table}
\arrayrulecolor{black}

\subsection{Zero-shot experiments}\label{sec:zero-shot}    

Based on its generative capabilities, TerraMind unlocks several zero-shot applications, like land-use segmentation, water body mapping, geo-localization, and vegetation mapping. In the following, we focus on water body mapping and geo-localization as image- and sequence-level zero-shot tasks.

\textbf{Water body mapping.} In Table~\ref{tab:all_sen1floods11_zeroshot_results}, we compare the zero-shot performance of TerraMind with its fine-tuned performance and other finetuned benchmarks for water body mapping. Overall, TerraMindv1-B achieves a zero-shot IoU of 45.4\% compared to SOTA-level fine-tuning performance of 82.2\% of DeCUR. In ablations with TerraMindv1-B-single trained on DynamicWorld LULC data, we boost this to up to 69.8\% suggesting that TerraMind harnesses up to over 80\% of the SOTA performance in zero-shot setting. Additionally, it's notable that none of the benchmarking model can be applied in a zero-shot context, highlighting the relevance of TerraMind's capabilities.

\begin{table}[h]
    \centering
    \small
    \setlength{\tabcolsep}{3pt}
    \begin{tabular}{@{} lccc @{}}
        \toprule
        \multicolumn{1}{l}{{Model}} & {Input} & Type & IoU$_{Water}$ \\
        \midrule
        \multicolumn{1}{l}{TerraMindv1-B}  & S-2 & zero-shot & 45.40  \\
        \multicolumn{1}{l}{TerraMindv1-B-single} & S-2 & zero-shot & 69.75 \\
        \multicolumn{2}{l}{Prithvi 2.0 / DeCUR / ...} & zero-shot & \textit{N/A} \\
        \midrule
        \multicolumn{1}{l}{Baseline \cite{Bonafilia_2020_CVPR_Workshops}}  & S-2 & finetune & 31.25  \\ 
        \multicolumn{1}{l}{Prithvi 2.0 300M} & S-2 & finetune & 80.97 \\ 
        \multicolumn{1}{l}{DeCUR} & S-2 & finetune & 82.17  \\ 
        \bottomrule
    \end{tabular}
    \caption{Zero-shot results of TerraMind on water body mapping compared to fine-tuned performance of benchmarks.}
    \label{tab:all_sen1floods11_zeroshot_results}
\end{table}

\begin{figure}[h]
    \centering
        \centering
        \includegraphics[width=0.8\linewidth]{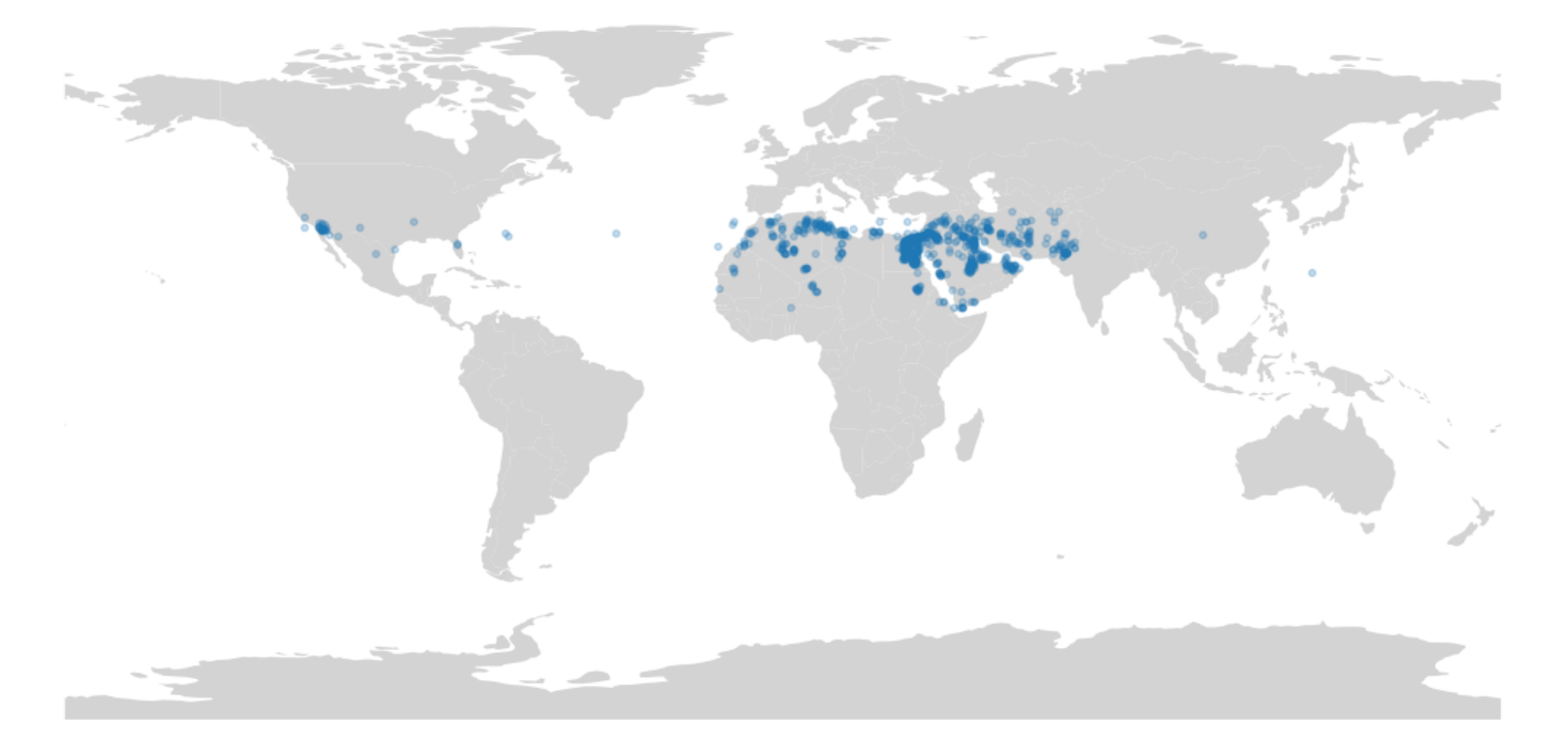}
        \caption{Prediction distribution of the land use class ``bare land'' with a sampling temperature of $T=1.0$ using TerraMindv1-B-single. TerraMind has an accurate internal representation of the geolocation of specific contexts, like land use classes.}
        \label{fig:lulc_geo_loc}
\end{figure}

\textbf{Geo-localization.} TerraMind is able to predict the geolocation of a specific data instance. To better visualize the geo-localization capabilities, we prompt the model for the most likely locations of the land use class ``bare land'' (deserts etc.) in a Monte-Carlo-sampling in Figure~\ref{fig:lulc_geo_loc}. The probability distribution of the model fits the expectation of where to find bare land, highlighting the Sahara region and middle-east, as well as Mexico and Southern California.

\subsection{Few-shot experiments}

TerraMind is trained via a cross-modal patch classification objective. Thus, we expect a well-structured latent space that clusters different concepts accurately. To investigate our hypothesis, we apply 1-Nearest-Neighbor (1-NN) classification experiments in the community-standard setting of 1-shot 5-way on two datasets: EuroSAT and METER-ML. In those experiments, there are no weight updates of any kind, so that we can assess the quality of the embedding space structure. In Table~\ref{tab:one-shot}, we observe that TerraMind outperforms several other benchmarks from both the CV and EO domain on the EuroSAT dataset by at least 10pp in accuracy. Our results further show that for methane source classification on METER-ML, TerraMind outperforms benchmark models and generalizes to high-resolution NAIP data with one order of magnitude higher resolution than the pre-training data. We present additional experiments with other few-shot settings in the supplementary material. 

\begin{table}[h]
\centering
\small
\begin{tabular}{llcc}
\toprule
Model & Input & EuroSAT & METER-ML \\
\midrule
{CLIP-ViT-B/16} & {S-2 RGB} & 57.00 & 29.15  \\
{CLIP-ViT-B/16} & {NAIP} & – & 32.01  \\
{DeCUR} & {S-2 L1C} & 50.54 & 27.87  \\
{Prithvi 1.0 100M} & {S-2 L1C} & 60.11 & 26.08\\
{Prithvi 2.0 300M} & {S-2 L1C}  & \underline{61.06} & 28.26  \\
\midrule
{{TerraMindv1-B}} & {S-2 L1C} & \textbf{70.83}  & \textbf{33.90}  \\
{{TerraMindv1-B}} & {NAIP} & –  & \underline{32.23}  \\
\bottomrule
\end{tabular}
\caption{1-shot 5-way classification results on EuroSAT and METER-ML measured in mean accuracy $\uparrow$, averaged over 200 runs. TerraMind outperforms benchmarks from CV and EO domain, suggesting a well-structured latent space.}
\label{tab:one-shot}
\end{table}

\subsection{Fine-tuning experiments}\label{sec:finetuning}

\begin{table*}[htb]
\centering
\small
\begin{adjustbox}{width=1\textwidth}
\begin{tabular}{lccccccccc|cc}
\toprule
Model & BurnSr$^\ast$ & MADOS$^\ast$ & PASTIS & Sen1Fl11 & FBP$^\ast$ & DEN$^\ast$ & CTM-SS & SN7$^\ast$ & AI4Farms$^\ast$ & Avg. mIoU & Avg. Rank \\
\midrule
CROMA                & 82.42        & 67.55  & 32.32 & \underline{90.89} & 51.83 & 38.29 & 49.38 & 59.28 & 25.65 & 55.29 & 6.61 \\
DOFA                 & 80.63        & 59.58  & 30.02 & 89.37            & 43.18 & \underline{39.29} & 51.33 & 61.84 & 27.07 & 53.59 & 8.22 \\
GFM-Swin             & 76.90        & 64.71  & 21.24 & 72.60            & 67.18 & 34.09 & 46.98 & 60.89 & 27.19 & 52.42 & 10.00 \\
Prithvi 1.0 100M     & 83.62        & 49.98  & 33.93 & 90.37            & 46.81 & 27.86 & 43.07 & 56.54 & 26.86 & 51.00 & 11.00 \\
RemoteCLIP           & 76.59        & 60.00  & 18.23 & 74.26            & \textbf{69.19} & 31.78 & 52.05 & 57.76 & 25.12 & 51.66 & 11.22 \\
SatlasNet            & 79.96        & 55.86  & 17.51 & 90.30            & 50.97 & 36.31 & 46.97 & 61.88 & 25.13 & 51.65 & 10.67 \\
Scale-MAE            & 76.68        & 57.32  & 24.55 & 74.13            & \underline{67.19} & 35.11 & 25.42 & \textbf{62.96} & 21.47 & 49.43 & 11.44 \\
SpectralGPT          & 80.47        & 57.99  & 35.44 & 89.07            & 33.42 & 37.85 & 46.95 & 58.86 & 26.75 & 51.87 & 10.11 \\
S.-S12-MoCo         & 81.58        & 51.76  & 34.49 & 89.26            & 53.02 & 35.44 & 48.58 & 57.64 & 25.38 & 53.02 & 10.06 \\
S.-S12-DINO         & 81.72        & 49.37  & 36.18 & 88.61            & 51.15 & 34.81 & 48.66 & 56.47 & 25.62 & 52.51 & 10.89 \\
S.-S12-MAE          & 81.91        & 49.90  & 32.03 & 87.79            & 51.92 & 34.08 & 45.80 & 57.13 & 24.69 & 51.69 & 12.39 \\
S.-S12-Data2Vec     & 81.91        & 44.36  & 34.32 & 88.15            & 48.82 & 35.90 & 54.03 & 58.23 & 24.23 & 52.22 & 10.72 \\  
\midrule 
UNet Baseline        & \textbf{84.51} & 54.79  & 31.60 & \textbf{91.42}   & 60.47 & \textbf{39.46} & 47.57 & \underline{62.09} & \textbf{46.34} & {57.58} & {4.89} \\
ViT Baseline         & 81.58        & 48.19  & 38.53 & 87.66            & 59.32 & 36.83 & 44.08 & 52.57 & \underline{38.37} & 54.13 & 10.28 \\
\midrule
TerraMindv1-B       & 82.42        & \underline{69.52} & 40.51 & 90.62            & 59.72 & 37.87 & \textbf{55.80} & 60.61 & 28.12 & \underline{58.35} & \underline{3.94} \\
TerraMindv1-L       & 82.93        & \textbf{75.57}  & \textbf{43.13} & 90.78            & 63.38 & 37.89     & \underline{55.04} & 59.98 & 27.47 & \textbf{59.57}    & \textbf{3.44} \\
\bottomrule
\end{tabular}
\end{adjustbox}
\caption{Performance evaluation of TerraMind using the PANGAEA evaluation protocol indicates higher mIoU values ($\uparrow$) and lower rank values ($\downarrow$). The best model per column is highlighted in bold, the second best is underscored. We indicate unimodal datasets with $^\ast$. Encoders are frozen for pretrained models, while U-Net and ViT baselines are trained from scratch for each specific task.}
\label{tab:pangaea-full-results}  
\end{table*}
Besides the novel capabilities that TerraMind introduces, we benchmark the fine-tuning performance of TerraMind in both unimodal and multimodal settings following the community-standard PANGAEA benchmark \cite{marsocci2024pangaea}. We summarize the results in Table~\ref{tab:pangaea-full-results}. Overall, TerraMindv1-B outperforms all other GeoFMs by at least 3pp avg. mIoU. Importantly, we observe that TerraMind is the only foundation model approach in EO that across the PANGAEA benchmark outperforms task-specific U-Net models. Performance increases by approximately 2pp avg. mIoU for TerraMindv1-L, with a peak of 5pp in multimodal datasets. Furthermore, TerraMindv1-L outperforms also specialised ViT baselines by 5pp avg. mIoU. 
Note that per suggestion of the PANGAEA authors, we exclude the xView2 and BioMassters task as we could not reproduce the reported performances. 
Finally, we assess the impact of leveraging multimodal data as input to TerraMindv1-B compared to utilizing either optical or radar data as unimodal input to better understand the effect of leveraging multimodal data in finetuning. We observe that across all three multimodal tasks, TerraMindv1-B performs best with access to both optical and radar data.

\begin{table}[h]
\centering
\small
\label{tab:multimodal}
\begin{tabular}{cccc}
\toprule
 & PASTIS         & Sen1Fl11       & CTM-SS        \\ \midrule
S-1         & 20.04          & 80.39          & 24.45          \\
S-2     & 40.20          & 89.57          & 50.90          \\\midrule
S-1 + S-2  & \textbf{40.51} & \textbf{90.62} & \textbf{55.80} \\ \bottomrule
\end{tabular}
\caption{Benefit of using multimodal input in the PANGAEA benchmark reported in mIoU (\%) $\uparrow$.}
\end{table}

\subsection{Thinking in modalities}  

We additionally evaluate the value of TiM tuning on water body mapping. We use S-1 or S-2 to generate artificial LULC data as additional input. Our results in Table~\ref{tab:sen1floods11_cot_results} indicate a superior performance of TiM tuning compared to leveraging uni-modal data by up to 2pp mIoU. This finding points us in the direction of TerraMind being able to generate data that improve downstream task performance. We provide additional results in the appendix.

\begin{table}[h] 
    \centering
    \small
    \begin{tabularx}{\linewidth}{lcCC}
        \toprule
        {Fine-Tuning} & {Input} & IoU$_{Water}$ & mIoU \\
        \midrule
        TerraMindv1-B  & S-1 & 68.00 & 81.06 \\
        TerraMindv1-B  & S-2 & 82.26 & 89.70 \\
        \midrule
        TerraMindv1-B TiM  & S-1 \textit{+ gen. LULC} & \textbf{72.25} & \textbf{83.65} \\
        TerraMindv1-B TiM  & S-2 \textit{+ gen. LULC} & \textbf{84.75}  & \textbf{91.14} \\
        \bottomrule
    \end{tabularx}
    \caption{Thinking-in-modalities (TiM) tuning compared with standard full fine-tuning approaches on the Sen1Floods11 dataset.}
    \label{tab:sen1floods11_cot_results}
\end{table}

\section{Conclusion}
\label{sec:conclusion}

TerraMind's approach of combining token-level and pixel-level data has unlocked a range of new model capabilities in EO. TerraMind demonstrates not only beyond state-of-the-art performance in community-standard benchmarks, it also represents the first fully generative multimodal model in the domain. Because of the ability of integrating heterogeneous data sources, we expect that TerraMind-like models will expand to multi-temporal, multi-resolution, and hyperspectral data to fully leverage the data rich ecosystem available in the Earth Observation domain.

\section{Acknowledgements}

This work is part of the Fostering Advancements in Foundation Models via Unsupervised and Self-supervised Learning for Downstream Tasks in Earth Observation (FAST-EO) project, which is funded by the European Space Agency (ESA) $\Phi$-lab under contract No. 4000143501/23/I-DT.
The authors gratefully acknowledge the Gauss Centre for Supercomputing e.V. (www.gauss-centre.eu) for funding this project by providing computing time on the GCS Supercomputer JUWELS \cite{juwels} at Jülich Supercomputing Centre (JSC).

{\small
\bibliographystyle{ieeenat_fullname}
\bibliography{11_references}

@String(CVPR= {IEEE Conf. Comput. Vis. Pattern Recog.})

@String(ICPR = {Int. Conf. Pattern Recog.})

@String(CVPRW= {IEEE Conf. Comput. Vis. Pattern Recog. Worksh.})

@String(CVPR  = {CVPR})

@String(ICPR  = {ICPR})

@String(CVPRW= {CVPRW})

@misc{mizrahi20234mmassivelymultimodalmasked,
      title={4M: Massively Multimodal Masked Modeling}, 
      author={David Mizrahi and Roman Bachmann and Oğuzhan Fatih Kar and Teresa Yeo and Mingfei Gao and Afshin Dehghan and Amir Zamir},
      year={2023},
      eprint={2312.06647},
      archivePrefix={arXiv},
      primaryClass={cs.CV},
      url={https://arxiv.org/abs/2312.06647}, 
}

@article{francis2024major,
  title={Major TOM: Expandable Datasets for Earth Observation},
  author={Francis, Alistair and Czerkawski, Mikolaj},
  journal={arXiv preprint arXiv:2402.12095},
  year={2024}
}

@article{wang2023ssl4eo,
  title={SSL4EO-S12: A large-scale multimodal, multitemporal dataset for self-supervised learning in Earth observation [Software and Data Sets]},
  author={Wang, Yi and Braham, Nassim Ait Ali and Xiong, Zhitong and Liu, Chenying and Albrecht, Conrad M and Zhu, Xiao Xiang},
  journal={IEEE Geoscience and Remote Sensing Magazine},
  volume={11},
  number={3},
  pages={98--106},
  year={2023},
  publisher={IEEE}
}

@misc{dem2022,
  title = {Copernicus DEM},
  howpublished = {\url{http://dx.doi.org/10.5270/ESA-c5d3d65}},
  DOI = {10.5270/esa-c5d3d65},
  journal = {Copernicus DEM},
  author = {European Space Agency},
  year = {2022}
}

@article{astruc2024anysat,
  title={AnySat: An Earth Observation Model for Any Resolutions, Scales, and Modalities},
  author={Astruc, Guillaume and Gonthier, Nicolas and Mallet, Clement and Landrieu, Loic},
  journal={arXiv preprint arXiv:2412.14123},
  year={2024}
}

@article{yu2024metaearth,
  title={Metaearth: A generative foundation model for global-scale remote sensing image generation},
  author={Yu, Zhiping and Liu, Chenyang and Liu, Liqin and Shi, Zhenwei and Zou, Zhengxia},
  journal={IEEE Transactions on Pattern Analysis and Machine Intelligence},
  year={2024},
  publisher={IEEE}
}

@article{van2017neural,
  title={Neural discrete representation learning},
  author={Van Den Oord, Aaron and Vinyals, Oriol and others},
  journal={Advances in neural information processing systems},
  volume={30},
  year={2017}
}

@article{mentzer2023finite,
  title={Finite scalar quantization: Vq-vae made simple},
  author={Mentzer, Fabian and Minnen, David and Agustsson, Eirikur and Tschannen, Michael},
  journal={arXiv preprint arXiv:2309.15505},
  year={2023}
}

@inproceedings{kenton2019bert,
  title={Bert: Pre-training of deep bidirectional transformers for language understanding},
  author={Kenton, Jacob Devlin Ming-Wei Chang and Toutanova, Lee Kristina},
  booktitle={Proceedings of naacL-HLT},
  volume={1},
  pages={2},
  year={2019},
  organization={Minneapolis, Minnesota}
}

@misc{dosovitskiy2021imageworth16x16words,
      title={An Image is Worth 16x16 Words: Transformers for Image Recognition at Scale}, 
      author={Alexey Dosovitskiy and Lucas Beyer and Alexander Kolesnikov and Dirk Weissenborn and Xiaohua Zhai and Thomas Unterthiner and Mostafa Dehghani and Matthias Minderer and Georg Heigold and Sylvain Gelly and Jakob Uszkoreit and Neil Houlsby},
      year={2021},
      eprint={2010.11929},
      archivePrefix={arXiv},
      primaryClass={cs.CV},
      url={https://arxiv.org/abs/2010.11929}, 
}

@misc{liu2024remoteclip,
      title={RemoteCLIP: A Vision Language Foundation Model for Remote Sensing}, 
      author={Fan Liu and Delong Chen and Zhangqingyun Guan and Xiaocong Zhou and Jiale Zhu and Qiaolin Ye and Liyong Fu and Jun Zhou},
      year={2024},
      eprint={2306.11029},
      archivePrefix={arXiv},
      primaryClass={cs.CV}
}

@ARTICLE{ChenweiWang,
  author={{Chenwei Wang, et al.}},   
  journal={IEEE Journal of Selected Topics in Applied Earth Observations and Remote Sensing (JSTARS), Vol. 15}, 
  title={{SAR Target Image Generation Method Using Azimuth-Controllable Generative Adversarial Network}},   
  year={2022. Online: \url{http://ieeexplore.ieee.org/stamp/stamp.jsp?tp=&arnumber=9933645&tag=1}}
}

@inproceedings{PrabhishekSingh,
author = {{Prabhishek Singh and Raj Shree}},
title = {{Analysis and effects of speckle noise in SAR images}},
year = {2016. DOI: 10.1109/ICACCAF.2016.7748978. Online: \url{http://ieeexplore.ieee.org/document/7748978}},
booktitle = {Proc. International Conference on Advances in Computing, Communication, \& Automation (ICACCA)},
}

@misc{XinyuBai,
      title={{Accelerating Diffusion for SAR-to-Optical Image Translation via Adversarial Consistency Distillation}},  
      author={{Xinyu Bai and Feng Xu}},
      year={2024. Online: \url{http://arxiv.org/pdf/2407.06095}},
      eprint={2407.06095},
      archivePrefix={arXiv},
      primaryClass={cs.CV}
}

@article{gomes2025lossy,
  title={Lossy Neural Compression for Geospatial Analytics: A Review},
  author={Gomes, Carlos and Wittmann, Isabelle and Robert, Damien and Jakubik, Johannes and Reichelt, Tim and Martone, Michele and Maurogiovanni, Stefano and Vinge, Rikard and Hurst, Jonas and Scheurer, Erik and others},
  journal={arXiv preprint arXiv:2503.01505},
  year={2025}
}

@ARTICLE{ZWang,
  author={{Z. Wang, A. C. Bovik, H. R. Sheikh, and E. P. Simoncelli}},   
  journal={IEEE Transactions on Image Processing, vol. 13, no. 4, pp. 600-612}, 
  title={{Image quality assessment: From error visibility to structural similarity}},   
  year={2004}
}

@inproceedings{AHore,
author = {{A. Hore and D. Ziou}},
title = {{Image quality metrics: PSNR vs. SSIM}},
year = {2010}, 
booktitle = {Proc. 20th International Conference on Pattern Recognition (ICPR), pp. 2366-2369},
}

@ARTICLE{KoheiArai,
  author={{Kohei Arai, Michihiro Mikamo, and Shunsuke Onishi}},    
  journal={International Journal of Advanced Computer Science and Applications (IJACSA), Vol. 14, No. 7}, 
  title={{Method for Image Quality Evaluation of Satellite-based SAR Data}},   
  year={2023. Online: \url{http://thesai.org/Downloads/Volume14No7/Paper_13-Method_for_Image_Quality_Evaluation_of_Satellite_based_SAR_Data.pdf}}
}

@inproceedings{visionpaper,
author = {Mai, Gengchen and Cundy, Chris and Choi, Kristy and Hu, Yingjie and Lao, Ni and Ermon, Stefano},
title = {Towards a Foundation Model for Geospatial Artificial Intelligence (Vision Paper)},
year = {2022},
isbn = {9781450395298},
publisher = {Association for Computing Machinery},
address = {New York, NY, USA},
url = {https://doi.org/10.1145/3557915.3561043},
doi = {10.1145/3557915.3561043},
booktitle = {Proceedings of the 30th International Conference on Advances in Geographic Information Systems},
articleno = {106},
numpages = {4},
location = {Seattle, Washington},
series = {SIGSPATIAL '22}
}

@inproceedings{clip,
  title={Learning transferable visual models from natural language supervision},
  author={Radford, Alec and Kim, Jong Wook and Hallacy, Chris and Ramesh, Aditya and Goh, Gabriel and Agarwal, Sandhini and Sastry, Girish and Askell, Amanda and Mishkin, Pamela and Clark, Jack and others},
  booktitle={International Conference on Machine Learning},
  pages={8748--8763},
  year={2021},
  organization={PmLR}
}

@article{mendieta2023gfm,
  title={GFM: Building Geospatial Foundation Models via Continual Pretraining},
  author={Mendieta, Matias and Han, Boran and Shi, Xingjian and Zhu, Yi and Chen, Chen and Li, Mu},
  journal={arXiv preprint arXiv:2302.04476},
  year={2023}
}

@inproceedings{
nascetti2023biomassters,
title={BioMassters: A Benchmark Dataset for Forest Biomass Estimation using Multi-modal Satellite Time-series},
author={Andrea Nascetti and RITU YADAV and Kirill Brodt and Qixun Qu and Hongwei Fan and Yuri Shendryk and Isha Shah and Christine Chung},
booktitle={Thirty-seventh Conference on Neural Information Processing Systems Datasets and Benchmarks Track},
year={2023},
url={https://openreview.net/forum?id=hrWsIC4Cmz}
}

@InProceedings{Bonafilia_2020_CVPR_Workshops,
author = {Bonafilia, Derrick and Tellman, Beth and Anderson, Tyler and Issenberg, Erica},
title = {Sen1Floods11: A Georeferenced Dataset to Train and Test Deep Learning Flood Algorithms for Sentinel-1},
booktitle = {Proceedings of the IEEE/CVF Conference on Computer Vision and Pattern Recognition (CVPR) Workshops},
month = {June},
year = {2020}
}

@article{dimitrovski2023current,
  title={Current trends in deep learning for Earth Observation: An open-source benchmark arena for image classification},
  author={Dimitrovski, Ivica and Kitanovski, Ivan and Kocev, Dragi and Simidjievski, Nikola},
  journal={ISPRS Journal of Photogrammetry and Remote Sensing},
  volume={197},
  pages={18--35},
  year={2023},
  publisher={Elsevier}
}

@article{yuan2021review,
  title={A review of deep learning methods for semantic segmentation of remote sensing imagery},
  author={Yuan, Xiaohui and Shi, Jianfang and Gu, Lichuan},
  journal={Expert Systems with Applications},
  volume={169},
  pages={114417},
  year={2021},
  publisher={Elsevier}
}

@article{wang2022self,
  title={Self-supervised learning in remote sensing: A review},
  author={Wang, Yi and Albrecht, Conrad M and Braham, Nassim Ait Ali and Mou, Lichao and Zhu, Xiao Xiang},
  journal={arXiv preprint arXiv:2206.13188},
  year={2022}
}

@article{zhang2024vision,
  title={Vision-language models for vision tasks: A survey},
  author={Zhang, Jingyi and Huang, Jiaxing and Jin, Sheng and Lu, Shijian},
  journal={IEEE Transactions on Pattern Analysis and Machine Intelligence},
  year={2024},
  publisher={IEEE}
}

@article{bordes2024introduction,
  title={An introduction to vision-language modeling},
  author={Bordes, Florian and Pang, Richard Yuanzhe and Ajay, Anurag and Li, Alexander C and Bardes, Adrien and Petryk, Suzanne and Ma{\~n}as, Oscar and Lin, Zhiqiu and Mahmoud, Anas and Jayaraman, Bargav and others},
  journal={arXiv preprint arXiv:2405.17247},
  year={2024}
}

@inproceedings{yang2024mma,
  title={Mma: Multi-modal adapter for vision-language models},
  author={Yang, Lingxiao and Zhang, Ru-Yuan and Wang, Yanchen and Xie, Xiaohua},
  booktitle={Proceedings of the IEEE/CVF Conference on Computer Vision and Pattern Recognition},
  pages={23826--23837},
  year={2024}
}

@inproceedings{ren2024timechat,
  title={Timechat: A time-sensitive multimodal large language model for long video understanding},
  author={Ren, Shuhuai and Yao, Linli and Li, Shicheng and Sun, Xu and Hou, Lu},
  booktitle={Proceedings of the IEEE/CVF Conference on Computer Vision and Pattern Recognition},
  pages={14313--14323},
  year={2024}
}

@inproceedings{han2024multimodal,
  title={Multimodal Large Language Models and Tunings: Vision, Language, Sensors, Audio, and Beyond},
  author={Han, Soyeon Caren and Cao, Feiqi and Poon, Josiah and Navigli, Roberto},
  booktitle={Proceedings of the 32nd ACM International Conference on Multimedia},
  pages={11294--11295},
  year={2024}
}

@inproceedings{jain2024vcoder,
  title={Vcoder: Versatile vision encoders for multimodal large language models},
  author={Jain, Jitesh and Yang, Jianwei and Shi, Humphrey},
  booktitle={Proceedings of the IEEE/CVF Conference on Computer Vision and Pattern Recognition},
  pages={27992--28002},
  year={2024}
}

@article{wu2025visionllm,
  title={Visionllm v2: An end-to-end generalist multimodal large language model for hundreds of vision-language tasks},
  author={Wu, Jiannan and Zhong, Muyan and Xing, Sen and Lai, Zeqiang and Liu, Zhaoyang and Chen, Zhe and Wang, Wenhai and Zhu, Xizhou and Lu, Lewei and Lu, Tong and others},
  journal={Advances in Neural Information Processing Systems},
  volume={37},
  pages={69925--69975},
  year={2025}
}

@inproceedings{lin2023multimodality,
  title={Multimodality helps unimodality: Cross-modal few-shot learning with multimodal models},
  author={Lin, Zhiqiu and Yu, Samuel and Kuang, Zhiyi and Pathak, Deepak and Ramanan, Deva},
  booktitle={Proceedings of the IEEE/CVF Conference on Computer Vision and Pattern Recognition},
  pages={19325--19337},
  year={2023}
}

@article{driess2023palm,
  title={Palm-e: An embodied multimodal language model},
  author={Driess, Danny and Xia, Fei and Sajjadi, Mehdi SM and Lynch, Corey and Chowdhery, Aakanksha and Wahid, Ayzaan and Tompson, Jonathan and Vuong, Quan and Yu, Tianhe and Huang, Wenlong and others},
  year={2023}
}

@article{fu2024video,
  title={Video-mme: The first-ever comprehensive evaluation benchmark of multi-modal llms in video analysis},
  author={Fu, Chaoyou and Dai, Yuhan and Luo, Yongdong and Li, Lei and Ren, Shuhuai and Zhang, Renrui and Wang, Zihan and Zhou, Chenyu and Shen, Yunhang and Zhang, Mengdan and others},
  journal={arXiv preprint arXiv:2405.21075},
  year={2024}
}

@inproceedings{cao2020behind,
  title={Behind the scene: Revealing the secrets of pre-trained vision-and-language models},
  author={Cao, Jize and Gan, Zhe and Cheng, Yu and Yu, Licheng and Chen, Yen-Chun and Liu, Jingjing},
  booktitle={Computer Vision--ECCV 2020: 16th European Conference, Glasgow, UK, August 23--28, 2020, Proceedings, Part VI 16},
  pages={565--580},
  year={2020},
  organization={Springer}
}

@inproceedings{cao2024maplm,
  title={Maplm: A real-world large-scale vision-language benchmark for map and traffic scene understanding},
  author={Cao, Xu and Zhou, Tong and Ma, Yunsheng and Ye, Wenqian and Cui, Can and Tang, Kun and Cao, Zhipeng and Liang, Kaizhao and Wang, Ziran and Rehg, James M and others},
  booktitle={Proceedings of the IEEE/CVF Conference on Computer Vision and Pattern Recognition},
  pages={21819--21830},
  year={2024}
}

@article{bommasani2021opportunities,
  title={On the opportunities and risks of foundation models},
  author={Bommasani, Rishi and Hudson, Drew A and Adeli, Ehsan and Altman, Russ and Arora, Simran and von Arx, Sydney and Bernstein, Michael S and Bohg, Jeannette and Bosselut, Antoine and Brunskill, Emma and others},
  journal={arXiv preprint arXiv:2108.07258},
  year={2021}
}

@inproceedings{manas2021seasonal,
  title={Seasonal contrast: Unsupervised pre-training from uncurated remote sensing data},
  author={Manas, Oscar and Lacoste, Alexandre and Gir{\'o}-i-Nieto, Xavier and Vazquez, David and Rodriguez, Pau},
  booktitle={Proceedings of the IEEE/CVF International Conference on Computer Vision},
  pages={9414--9423},
  year={2021}
}

@article{li2020object,
  title={Object detection in optical remote sensing images: A survey and a new benchmark},
  author={Li, Ke and Wan, Gang and Cheng, Gong and Meng, Liqiu and Han, Junwei},
  journal={ISPRS journal of photogrammetry and remote sensing},
  volume={159},
  pages={296--307},
  year={2020},
  publisher={Elsevier}
}

@article{zhu2017deep,
  title={Deep learning in remote sensing: A comprehensive review and list of resources},
  author={Zhu, Xiao Xiang and Tuia, Devis and Mou, Lichao and Xia, Gui-Song and Zhang, Liangpei and Xu, Feng and Fraundorfer, Friedrich},
  journal={IEEE geoscience and remote sensing magazine},
  volume={5},
  number={4},
  pages={8--36},
  year={2017},
  publisher={IEEE}
}

@article{chen2021self,
  title={Self-supervised Change Detection in Multi-view Remote Sensing Images},
  author={Chen, Yuxing and Bruzzone, Lorenzo},
  journal={arXiv preprint arXiv:2103.05969},
  year={2021}
}

@misc{xiong2024neural,
      title={Neural Plasticity-Inspired Foundation Model for Observing the Earth Crossing Modalities}, 
      author={Zhitong Xiong and Yi Wang and Fahong Zhang and Adam J. Stewart and Joëlle Hanna and Damian Borth and Ioannis Papoutsis and Bertrand Le Saux and Gustau Camps-Valls and Xiao Xiang Zhu},
      year={2024},
      eprint={2403.15356},
      archivePrefix={arXiv},
      primaryClass={cs.CV}
}

@misc{fibaek2024phileo,
      title={{PhilEO Bench: Evaluating Geo-Spatial Foundation Models}}, 
      author={Casper Fibaek and Luke Camilleri and Andreas Luyts and Nikolaos Dionelis and Bertrand Le Saux},
      year={In Proc. Int Geoscience and Remote Sensing Symposium (IGARSS), 2024},
      eprint={2401.04464},
      archivePrefix={arXiv},
      primaryClass={cs.CV}
}

@misc{jakubik2023foundation,
      title={Foundation Models for Generalist Geospatial Artificial Intelligence}, 
      author={Johannes Jakubik and Sujit Roy and C. E. Phillips and Paolo Fraccaro and Denys Godwin and Bianca Zadrozny and Daniela Szwarcman and Carlos Gomes and Gabby Nyirjesy and Blair Edwards and Daiki Kimura and Naomi Simumba and Linsong Chu and S. Karthik Mukkavilli and Devyani Lambhate and Kamal Das and Ranjini Bangalore and Dario Oliveira and Michal Muszynski and Kumar Ankur and Muthukumaran Ramasubramanian and Iksha Gurung and Sam Khallaghi and Hanxi and Li and Michael Cecil and Maryam Ahmadi and Fatemeh Kordi and Hamed Alemohammad and Manil Maskey and Raghu Ganti and Kommy Weldemariam and Rahul Ramachandran},
      year={2023},
      eprint={2310.18660},
      archivePrefix={arXiv},
      primaryClass={cs.CV}
}

@misc{han2024bridging,
      title={Bridging Remote Sensors with Multisensor Geospatial Foundation Models}, 
      author={Boran Han and Shuai Zhang and Xingjian Shi and Markus Reichstein},
      year={2024},
      eprint={2404.01260},
      archivePrefix={arXiv},
      primaryClass={cs.CV}
}

@misc{tuia2023artificial,
      title={Artificial intelligence to advance Earth observation: a perspective}, 
      author={Devis Tuia and Konrad Schindler and Begüm Demir and Gustau Camps-Valls and Xiao Xiang Zhu and Mrinalini Kochupillai and Sašo Džeroski and Jan N. van Rijn and Holger H. Hoos and Fabio Del Frate and Mihai Datcu and Jorge-Arnulfo Quiané-Ruiz and Volker Markl and Bertrand Le Saux and Rochelle Schneider},
      year={2023},
      eprint={2305.08413},
      archivePrefix={arXiv},
      primaryClass={cs.CV}
}

@misc{li2024visionlanguage,
      title={Vision-Language Models in Remote Sensing: Current Progress and Future Trends}, 
      author={Xiang Li and Congcong Wen and Yuan Hu and Zhenghang Yuan and Xiao Xiang Zhu},
      year={2024},
      eprint={2305.05726},
      archivePrefix={arXiv},
      primaryClass={cs.CV}
}

@article{terramesh,
  title={TerraMesh: A Planetary Mosaic of Multimodal Earth Observation Data},
  author={Blumenstiel, Benedikt and Fraccaro, Paolo and Marsocci, Valerio and Jakubik, Johannes and Maurogiovanni, Stefano and Czerkawski, Mikolaj and Sedona, Rocco and Cavallaro, Gabriele and Brunschwiler, Thomas and Bernabe-Moreno, Juan and Longépé, Nicolas},
  journal={arXiv preprint arXiv:2504.11172},
  year={2025}
}

@article{juwels,
  title={JUWELS cluster and booster: Exascale pathfinder with modular supercomputing architecture at Juelich Supercomputing Centre},
  author={Alvarez, Damian},
  journal={Journal of large-scale research facilities JLSRF},
  volume={7},
  pages={A183--A183},
  year={2021}
}

@article{yang2024multi,
  title={Multi-modal graph neural networks for localized off-grid weather forecasting},
  author={Yang, Qidong and Giezendanner, Jonathan and Civitarese, Daniel Salles and Jakubik, Johannes and Schmitt, Eric and Chandra, Anirban and Vila, Jeremy and Hohl, Detlef and Hill, Chris and Watson, Campbell and others},
  journal={arXiv preprint arXiv:2410.12938},
  year={2024}
}

@article{marimo2025beyond,
  title={Beyond the Visible: Multispectral Vision-Language Learning for Earth Observation},
  author={Marimo, Clive Tinashe and Blumenstiel, Benedikt and Nitsche, Maximilian and Jakubik, Johannes and Brunschwiler, Thomas},
  journal={arXiv preprint arXiv:2503.15969},
  year={2025}
}

@misc{marsocci2024crosssensor,
      title={Cross-sensor self-supervised training and alignment for remote sensing}, 
      author={Valerio Marsocci and Nicolas Audebert},
      year={2024},
      eprint={2405.09922},
      archivePrefix={arXiv},
      primaryClass={cs.CV}
}

@misc{fuller2023croma,
      title={CROMA: Remote Sensing Representations with Contrastive Radar-Optical Masked Autoencoders}, 
      author={Anthony Fuller and Koreen Millard and James R. Green},
      year={2023},
      eprint={2311.00566},
      archivePrefix={arXiv},
      primaryClass={cs.CV}
}

@article{swope2021representation,
  title={Representation learning for remote sensing: An unsupervised sensor fusion approach},
  author={Swope, Aidan M and Rudelis, Xander H and Story, Kyle T},
  journal={arXiv preprint arXiv:2108.05094},
  year={2021}
}

@article{machado2020airound,
  title={AiRound and CV-BrCT: Novel multiview datasets for scene classification},
  author={Machado, Gabriel and Ferreira, Edemir and Nogueira, Keiller and Oliveira, Hugo and Brito, Matheus and Gama, Pedro Henrique Targino and dos Santos, Jefersson Alex},
  journal={IEEE Journal of Selected Topics in Applied Earth Observations and Remote Sensing},
  volume={14},
  pages={488--503},
  year={2020},
  publisher={IEEE}
}

@article{deuser2023sample4geo,
  title={Sample4Geo: Hard Negative Sampling For Cross-View Geo-Localisation},
  author={Deuser, Fabian and Habel, Konrad and Oswald, Norbert},
  journal={arXiv preprint arXiv:2303.11851},
  year={2023}
}

@inproceedings{audebert_joint_2017,
  title = {Joint {{Learning}} from {{Earth Observation}} and {{OpenStreetMap Data}} to {{Get Faster Better Semantic Maps}}},
  booktitle = {Proceedings of the {{IEEE Conference}} on {{Computer Vision}} and {{Pattern Recognition Workshops}} ({{CVPRW}})},
  author = {Audebert, Nicolas and Le Saux, Bertrand and Lefèvre, Sébastien},
  date = {2017-07},
  pages = {1552--1560},
  location = {{Honolulu, United States}},
  doi = {10.1109/CVPRW.2017.199},
  year = {2017}
}

@article{hafner2022unsupervised,
  title={Unsupervised domain adaptation for global urban extraction using sentinel-1 SAR and sentinel-2 MSI data},
  author={Hafner, Sebastian and Ban, Yifang and Nascetti, Andrea},
  journal={Remote Sensing of Environment},
  volume={280},
  pages={113192},
  year={2022},
  publisher={Elsevier}
}

@article{li2022deep,
  title={Deep learning in multimodal remote sensing data fusion: A comprehensive review},
  author={Li, Jiaxin and Hong, Danfeng and Gao, Lianru and Yao, Jing and Zheng, Ke and Zhang, Bing and Chanussot, Jocelyn},
  journal={International Journal of Applied Earth Observation and Geoinformation},
  volume={112},
  pages={102926},
  year={2022},
  publisher={Elsevier}
}

@article{marsocci2024pangaea,
  title={PANGAEA: A global and inclusive benchmark for geospatial foundation models},
  author={Marsocci, Valerio and Jia, Yuru and Bellier, Georges Le and Kerekes, David and Zeng, Liang and Hafner, Sebastian and Gerard, Sebastian and Brune, Eric and Yadav, Ritu and Shibli, Ali and others},
  journal={arXiv preprint arXiv:2412.04204},
  year={2024}
}

@article{zhao2022cnn,
  title={CNN, RNN, or ViT? An evaluation of different deep learning architectures for spatio-temporal representation of sentinel time series},
  author={Zhao, Linying and Ji, Shunping},
  journal={IEEE Journal of Selected Topics in Applied Earth Observations and Remote Sensing},
  volume={16},
  pages={44--56},
  year={2022},
  publisher={IEEE}
}

@article{shafique2022deep,
  title={Deep learning-based change detection in remote sensing images: A review},
  author={Shafique, Ayesha and Cao, Guo and Khan, Zia and Asad, Muhammad and Aslam, Muhammad},
  journal={Remote Sensing},
  volume={14},
  number={4},
  pages={871},
  year={2022},
  publisher={MDPI}
}

@article{lahrichi2025self,
  title={Is Self-Supervised Pre-training on Satellite Imagery Better than ImageNet? A Systematic Study with Sentinel-2},
  author={Lahrichi, Saad and Sheng, Zion and Xia, Shufan and Bradbury, Kyle and Malof, Jordan},
  journal={arXiv preprint arXiv:2502.10669},
  year={2025}
}

@misc{khanna2023diffusionsat,
      title={DiffusionSat: A Generative Foundation Model for Satellite Imagery}, 
      author={Samar Khanna and Patrick Liu and Linqi Zhou and Chenlin Meng and Robin Rombach and Marshall Burke and David Lobell and Stefano Ermon},
      year={2023},
      eprint={2312.03606},
      archivePrefix={arXiv},
      primaryClass={cs.CV}
}

@misc{vanetten2019spacenet,
      title={SpaceNet: A Remote Sensing Dataset and Challenge Series}, 
      author={Adam Van Etten and Dave Lindenbaum and Todd M. Bacastow},
      year={2019},
      eprint={1807.01232},
      archivePrefix={arXiv},
      primaryClass={cs.CV}
}

@inproceedings{
garioud2023flair,
title={{FLAIR} : a Country-Scale Land Cover Semantic Segmentation Dataset From Multi-Source Optical Imagery},
author={Anatol Garioud and Nicolas Gonthier and Loic Landrieu and Apolline De Wit and Marion Valette and Marc Poup{\'e}e and Sebastien Giordano and Boris Wattrelos},
booktitle={Thirty-seventh Conference on Neural Information Processing Systems Datasets and Benchmarks Track},
year={2023},
url={https://openreview.net/forum?id=LegGqdch92}
}

@misc{astruc2024omnisat,
      title={OmniSat: Self-Supervised Modality Fusion for Earth Observation}, 
      author={Guillaume Astruc and Nicolas Gonthier and Clement Mallet and Loic Landrieu},
      year={2024},
      eprint={2404.08351},
      archivePrefix={arXiv},
      primaryClass={cs.CV}
}

@online{durnov_xview2_2020,
	title = {xview2 1st place solution},
	url = {https://github.com/vdurnov/xview2_1st_place_solution},
	author = {Durnov, Victor},
	urldate = {2022-01-10},
	date = {2020-02-27},
}

@article{nedungadi2024mmearth,
  title={MMEarth: Exploring multi-modal pretext tasks for geospatial representation learning},
  author={Nedungadi, Vishal and Kariryaa, Ankit and Oehmcke, Stefan and Belongie, Serge and Igel, Christian and Lang, Nico},
  journal={arXiv preprint arXiv:2405.02771},
  year={2024}
}

@misc{paolo2022xview3sardetectingdarkfishing,
      title={xView3-SAR: Detecting Dark Fishing Activity Using Synthetic Aperture Radar Imagery}, 
      author={Fernando Paolo and Tsu-ting Tim Lin and Ritwik Gupta and Bryce Goodman and Nirav Patel and Daniel Kuster and David Kroodsma and Jared Dunnmon},
      year={2022},
      eprint={2206.00897},
      archivePrefix={arXiv},
      primaryClass={cs.CV},
      url={https://arxiv.org/abs/2206.00897}, 
}

@article{francis2024sensor,
  title={Sensor independent cloud and shadow masking with partial labels and multimodal inputs},
  author={Francis, Alistair},
  journal={IEEE Transactions on Geoscience and Remote Sensing},
  year={2024},
  publisher={IEEE}
}

@misc{liu2022convnet2020s,
      title={A ConvNet for the 2020s}, 
      author={Zhuang Liu and Hanzi Mao and Chao-Yuan Wu and Christoph Feichtenhofer and Trevor Darrell and Saining Xie},
      year={2022},
      eprint={2201.03545},
      archivePrefix={arXiv},
      primaryClass={cs.CV},
      url={https://arxiv.org/abs/2201.03545}, 
}

@misc{li2024llavanext,
    title={LLaVA-NeXT: Stronger LLMs Supercharge Multimodal Capabilities in the Wild},
    url={https://llava-vl.github.io/blog/2024-05-10-llava-next-stronger-llms/},
    author={Li, Bo and Zhang, Kaichen and Zhang, Hao and Guo, Dong and Zhang, Renrui and Li, Feng and Zhang, Yuanhan and Liu, Ziwei and Li, Chunyuan},
    month={May},
    year={2024}
}

@article{blumenstiel2025ssl4eos12v11,
  title={{SSL4EOS12 v1.1 – A Multimodal, Multiseasonal Dataset for Pretraining}},
  author={Blumenstiel, Benedikt and Braham, Nassim Ait Ali and Albrecht, Conrad M and Maurogiovanni, Stefano and Fraccaro, Paolo},
  journal={arXiv preprint arXiv:2503.00168},
  year={2025}
}

@article{snell2017prototypical,
  title={Prototypical networks for few-shot learning},
  author={Snell, Jake and Swersky, Kevin and Zemel, Richard},
  journal={Advances in neural information processing systems},
  volume={30},
  year={2017}
}
}

\clearpage
\maketitlesupplementary


In the following, we provide additional information on our data, the pretraining of TerraMind and its tokenizers, the quality of the tokenization, any-to-any generation matrices, and comparisons of TerraMind in unimodal and multimodal finetuning against specialized U-Net and ViT models.

\section{TerraMesh Dataset}     

\begin{figure*}[hbt!]
    \centering
    \includegraphics[width=\textwidth]{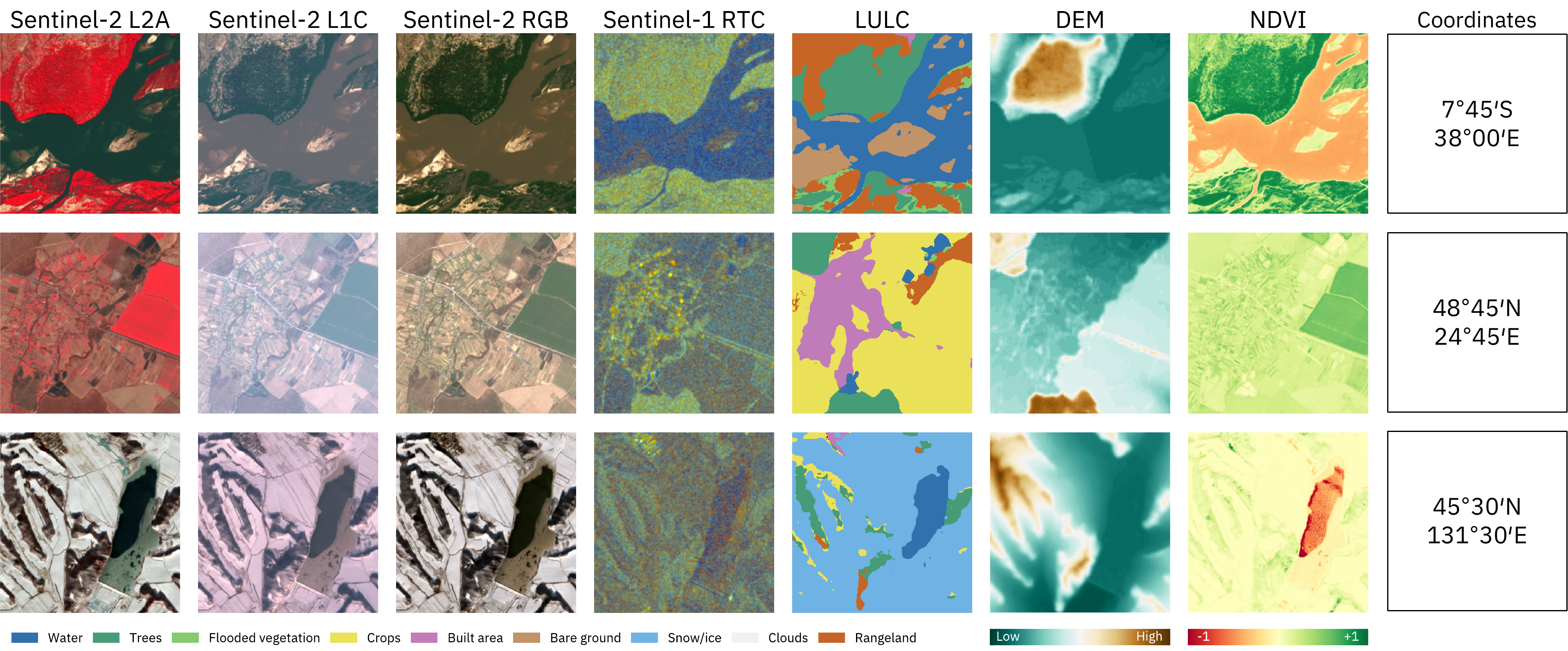}
    \caption{Visualization of the spatial-temporal alignment across modalities in TerraMesh. S-2 L2A uses IRRG pseudo-coloring and S-1 RTC is visualized in db scale as VH-VV-VV/VH. Copernicus DEM is scaled based on the image value range.}
    \label{fig:terramesh}
\end{figure*}

All versions of TerraMind have been pretrained on TerraMesh or a subset of it. TerraMesh is a comprehensive multimodal Earth observation dataset designed for large-scale model pre-training. It will be made publicly available under a permissive license in a preprint during the review process of this paper. The dataset includes nine modalities and we visualize examples of the dataset in Figure~\ref{fig:terramesh}.

The dataset contains over 9 million globally distributed, spatiotemporally aligned samples across nine core modalities. Each modality is precisely co-registered at a 10-meter resolution, primarily based on Sentinel-2 grids. The S-1 and S-2 samples are sourced from MajorTOM-Core~\cite{francis2024major} and SSL4EO-S12\,v1.1~\cite{blumenstiel2025ssl4eos12v11}.
It integrates Sentinel-1 SAR data with Sentinel-2 optical data (L1C top-of-atmosphere and L2A bottom-of-atmosphere reflectance), ensuring versatility for various downstream tasks. Because the source datasets contain only one S-1 product, each sample has either S-1 GRD or S-1 RTC data. Additionally, TerraMesh includes normalized difference vegetation index (NDVI) maps derived from Sentinel-2, Copernicus digital elevation model (DEM) data providing topographic context, and land-use/land-cover (LULC) maps from ESRI, enhanced with accurate cloud masks generated by the SEnSeI\,v2 model\cite{francis2024sensor}.

To ensure broad geographic and thematic diversity, TerraMesh employs subsampling techniques, selectively including representative samples from each global ecoregion and land-cover class, while downsampling highly homogeneous regions such as deserts and tundra. Another critical aspect is the data preprocessing pipeline, which includes reprojection, temporal alignment, and filtering to minimize missing data and artifacts, ensuring high-quality, analysis-ready samples.

TerraMind.v1-B-single was pre-trained on a subset of TerraMesh with one million samples, specifically the SSL4EOS12\,v1.1 locations, using only four image modalities: S-2 L2A, S-1 GRD, DEM, and LULC. Additionally, we performed continuous pre-training with image captions. These captions were created using LLaVA-Next~\cite{li2024llavanext} and Overture Maps data \cite{marimo2025beyond}. The automated captioning pipeline includes a prompt with a chain-of-thought process to generate diverse captions. The captioning model is asked to generate three question-answer pairs and describe the full image later. We use the S-2 RGB bands and Overture base layer tags as inputs.
Domain experts evaluated a subset of 1.3k captions, resulting in 69\% of the captions without any hallucinations while the average completeness scores were 3.87 on a scale from 0 to 5. 

\section{Pretraining details}

In this section, we give additional details on the pretraining of both TerraMind and its tokenizers.

\subsection{Tokenizer models}

\color{black}
The tokenizer models are pretrained using a Vision Transformer (ViT) encoder and a patched UNet decoder, with input images ranging from 224x224 to 256x256 in size. The model was trained with patch sizes of 16x16 for the ViT encoder and 4x4 for the UNet decoder. A tanh MLP was used before the quantizer, as outlined in the ViT-VQGAN paper, to enhance tokenization quality.

The model utilized a Finite-Scalar Quantization (FSQ) approach with a codebook size of 8-8-8-6-5, aiming to learn consistent and abstract representations across image patches. The latent dimension was set to 5. We leverage the normalization of codebook entries to the unit sphere during training. This concept is borrowed from the ViT-VQGAN approach, which applies a specific form of normalization to improve the quality and efficiency of learned representations. Additionally, an EMA-based quantizer was used with a decay rate of 0.99 to track and improve quantization over time.

During diffusion-based pretraining, the model was trained for 1000 timesteps using a linear beta schedule, with MSE loss as the objective. The training leveraged half-precision (fp16) and used an AdamW optimizer with specific learning rate scheduling and warmup strategies. The model also incorporated model EMA for stable training and set a batch size of 1 per GPU with various regularization techniques like grad clipping and random horizontal flips.
\normalcolor

We pretrained the TerraMind tokenizers for image-like modalities with DDP on 4 GPUs for a total of 100 epochs on the respective modality of TerraMesh. We use a base learning rate of 1e-4, an effective batch size of 64 samples per GPU, i.e. the global batch size is 256. We reach a GPU utilization of 99\% for single channel modalities like LULC and NDVI, and over 80\% for all multi-channel modalities.

\subsection{TerraMind}

We pretrained both TerraMindv1-B and TerraMindv1-L with DDP on 32 GPUs. We determine the global batch size based on initial experimental runs comparing a global batch size of 2K, 4K, and 8K. In addition, we determine the base learning rate starting from 1e-4 and iteratively experimented with half and double learning rates. Ultimately, we end up with a base learning rate of 2e-4 for a cosine annealing scheduler set to run for 500B tokens. For the v1-L model, we reach a GPU utilization of 85+\%. Overall, the training of TerraMindv1-B took 12 days on 32 A100 GPUs, i.e., 9'216 GPU hours. Over the course of the pretraining, we also experiment with different configurations of the Dirichlet sampling distribution. In total, the pretraining experiments have been approximately three times larger than the final runs resulting in approximately 30K GPU hours allocated for pretraining.

\begin{figure}[h]
        \centering
        \includegraphics[width=0.8\linewidth]{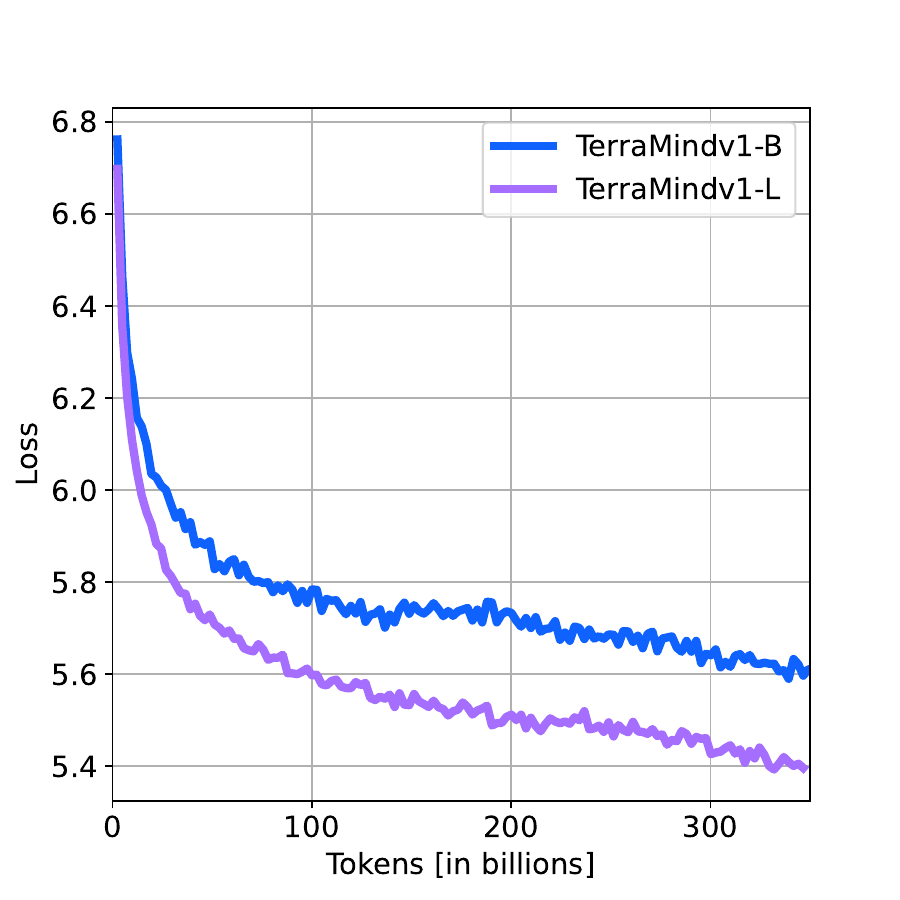}
        \caption{Example of the scaling behavior of TerraMind comparing v1-B and v1-L models for the first 350B tokens on the validation loss of optical S-2 L2A data. Overall, TerraMind-L outperforms TerraMind-B after approximately 10\% of the training schedule of the large model.}
        \label{fig:scaling}
\end{figure}

We provide an overview on the scaling dynamics when going from TerraMindv1-B to TerraMind v1-L in Figure~\ref{fig:scaling} with identical hyperparameters and compute. Overall, as expected, we observe a significant gap in the validation losses across modalities. We finally provide the validation losses per modality after pretraining of TerraMindv1-B and TerraMindv1-L in Table~\ref{tab:val_losses}.

\begin{table}[htb]
    \centering
    \small
    \begin{tabular}{lccccc}
        \toprule
        Model & S-2 L2A & S-1 GRD & S-1 RTC & DEM & NDVI \\
        \midrule
        Random & 9.68 & 9.68 & 9.68 & 9.68 & 9.68 \\
        \midrule
        V1-B & 5.67 & 7.84 & 7.64 & 2.19 & 6.42 \\
        V1-L & \textbf{5.34} & \textbf{7.69} & \textbf{7.53} & \textbf{2.14} & \textbf{6.25} \\
        \bottomrule 
    \end{tabular}
    \caption{Validation losses of full pre-training of TerraMindv1-B and v1-L.}
    \label{tab:val_losses}   
\end{table}

\section{Tokenizer performance and general learnings}     
In the following, we provide details on the tokenizations of TerraMind. At least for image-like modalities, the tokenizations represent an important and computationally heavy phase of the pretraining, which is why we highlight important learnings in the following. 

\textbf{Learnings.} Overall, we learned that the tokenizer performance can be quite sensitive, which is especially related to the significant bottleneck compression of up to 3000x after the encoder. When leveraging finite-scalar quantization (FSQ) instead of vector quantization (VQ), we observed exactly what the original FSQ paper \cite{mentzer2023finite} claims: FSQ makes quantization easier -- yet in our experiments it did not improve the reconstruction performance in terms of MSE losses. We leverage FSQ as the training was more stable and less sensitive to the learning rate, which is likely related to the fact that, unlike VQ, FSQ does not require an additional codebook loss. We still observed that all tokenizer models were sensitive to the learning rate, with higher learning rates resulting in non-differentiability (NaN losses), and low learning rates caused blurry results. 

In addition, we experimented with the codebook size. In our experiments, we observed that the level of detail in the reconstructions was significantly higher for single channel input compared to multi channel input (e.g., 12 band S2-L2A data). Naturally, with less channels, the compression bottleneck for equal-sized codebooks is lower. Therefore, we hypothesized whether multi-spectral data requires larger codebook sizes to obtain higher level of detail in the reconstructions. In contrast to our expectation, when increasing the codebook size over 16K for modalities with more than three input channels, the reconstructions had significant artefacts. This suggests that even though the compression bottleneck is lower, higher codebook sizes are more difficult for the model to use, which is in line with previous literature. However, we were surprised to see more artefacts in the reconstructions of models with a codebook size 32K compared to 16K. 

Finally, we experimented with exponential moving average (EMA) updates for the tokenizer models. As expected, the models were less responsive to gradient updates. The resulting reconstructions smoothed out more of finegrained features. Together with the generative diffusion process in the tokenizer decoder, the resulting reconstructions often looked like hallucinations, e.g. bridges over rivers were not existing anymore in the reconstruction images. We therefore decided to ommit expotential moving average in our tokenizer models.

\color{black}
\subsection{FSQ vs. VQ}

\begin{figure}[h]
    \centering
    \includegraphics[width=\linewidth]{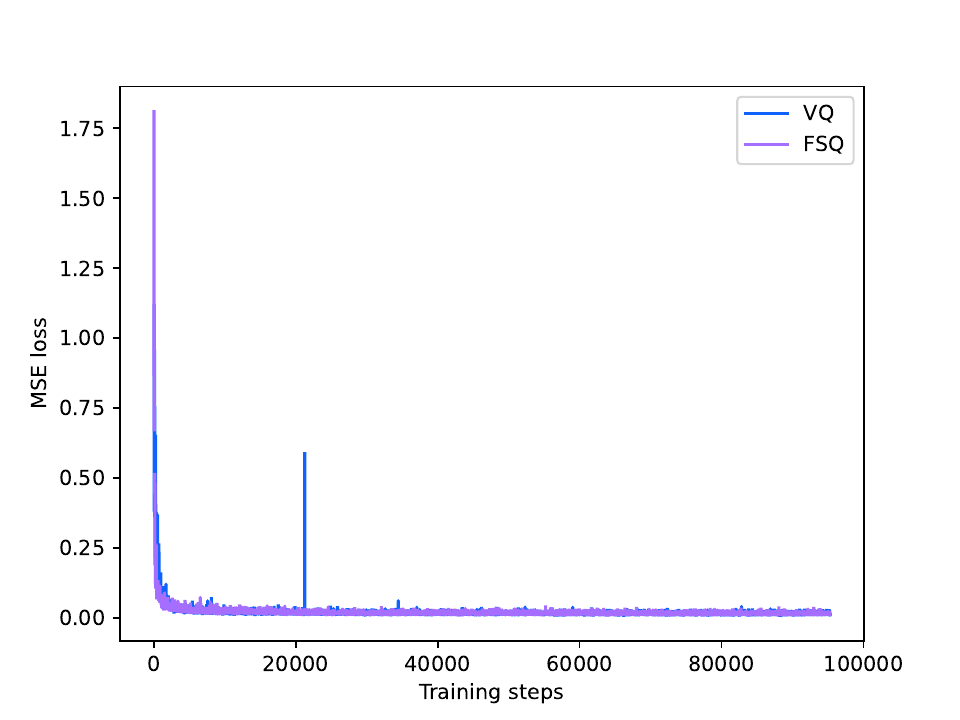}
    \caption{\textcolor{black}{Pretraining reconstruction losses of S-2 L2A modality comparing finite-scalar quantization (FSQ) and vector quantization (VQ) approaches. Overall, both approaches reach the same level of performance. The FSQ approach converges smoother than VQ, while requiring less tuning.}}
    \label{fig:fsq_loss}
\end{figure}

\begin{figure}[h]
    \centering
    \includegraphics[width=\linewidth]{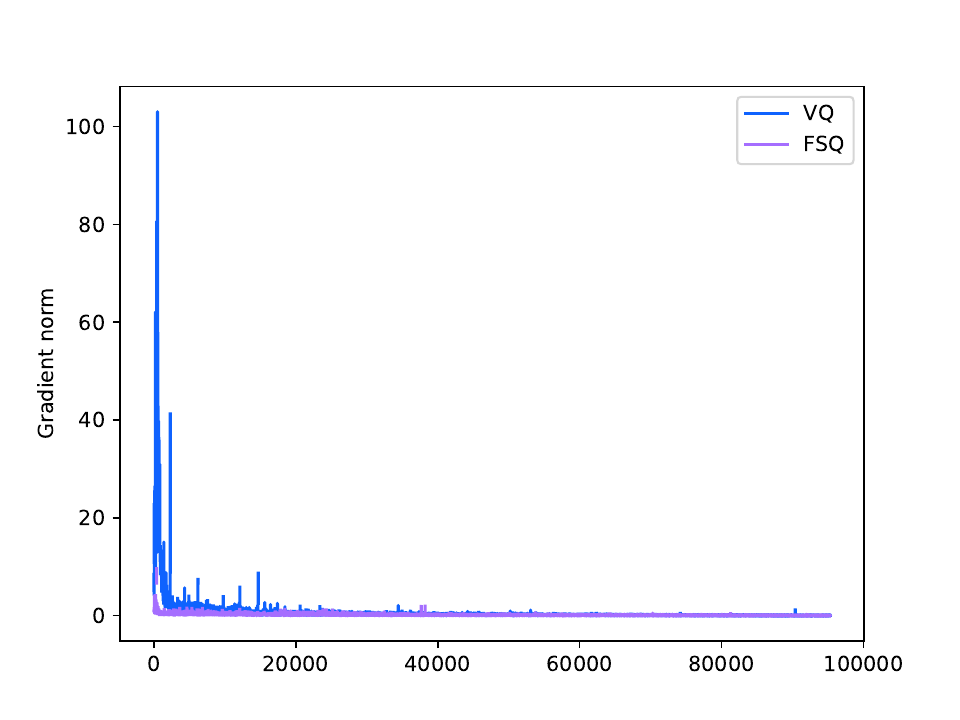}
    \caption{\textcolor{black}{Gradient norms for pretraining of S-2 L2A tokenizers comparing finite-scalar quantization (FSQ) and vector quantization (VQ) approaches. The FSQ approach converges smoother than VQ, while requiring less tuning.}}
    \label{fig:fsq_grads}
\end{figure}

Generally, our pretraining experiments comparing FSQ with vector quantization suggest that both approaches can achieve the same level of performance, yet reaching optimal levels of performance with VQ is regarded to be more challenging than using FSQ. We visualize this through (a) the reconstruction loss and (b) the gradient norms of the tokenizer pretraining on S-2 L2A data in Figures~\ref{fig:fsq_loss} and \ref{fig:fsq_grads}, respectively. Overall, we observe that both approaches reach the same level of convergence, however FSQ requires less tuning and is generally more stable than VQ. This especially also applies for the grad norms.

\normalcolor
\normalcolor

\textbf{Performance.}    
In the following, we assess the accuracy of our tokenizer models. Besides visual quality assessments and quantitative assessments with MSE metrics, we were particularly interested in whether our tokenizers exhibit geospatial biases. Understanding this is crucial to ensure TerraMind has a uniform level of performance across the globe. In addition, we investigate the reconstructions of radar data in more detail, as radar data by nature includes significant noise in the amplitude data. This could interfere with the noise generation in the diffusion process of the decoder, which is why we assess the structure of the reconstructions using SSIM and PSNR metrics.

\begin{figure}
    \centering
    \includegraphics[width=\linewidth]{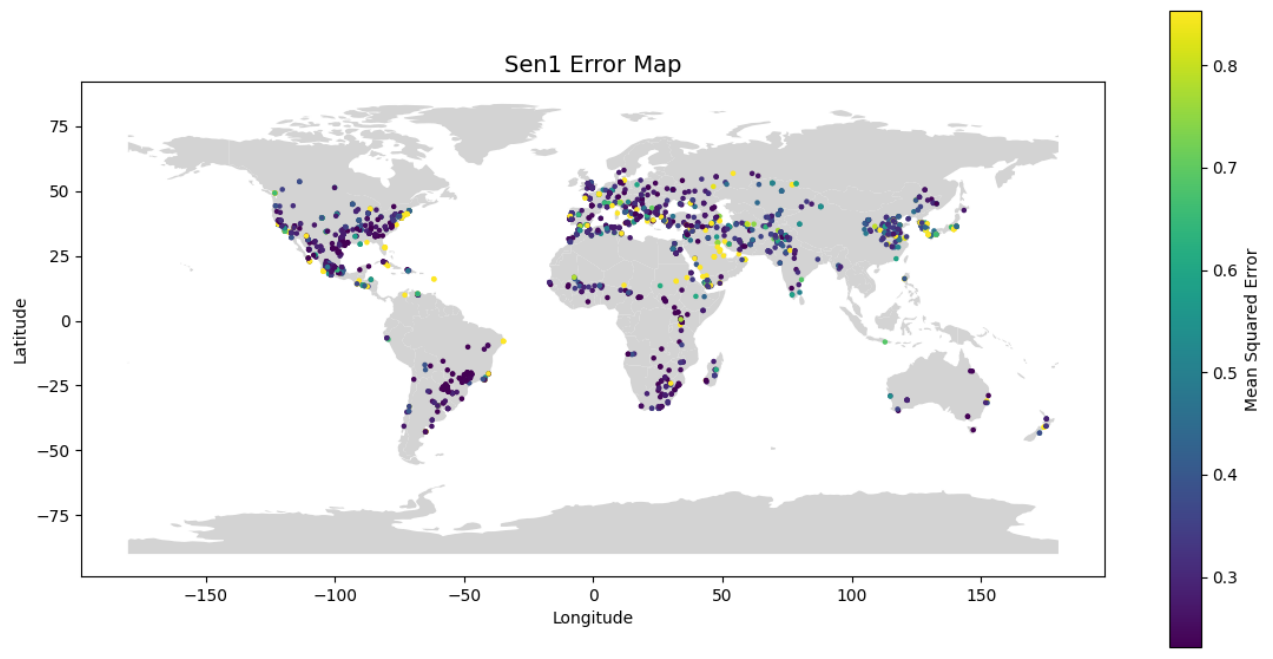}
    \caption{Spatial distribution of mean squared errors of the S-1 tokenizer on the validation set of the pretraining data.}
    \label{fig:s1-error-map}
\end{figure}

\begin{figure} 
    \centering
    \includegraphics[width=\linewidth]{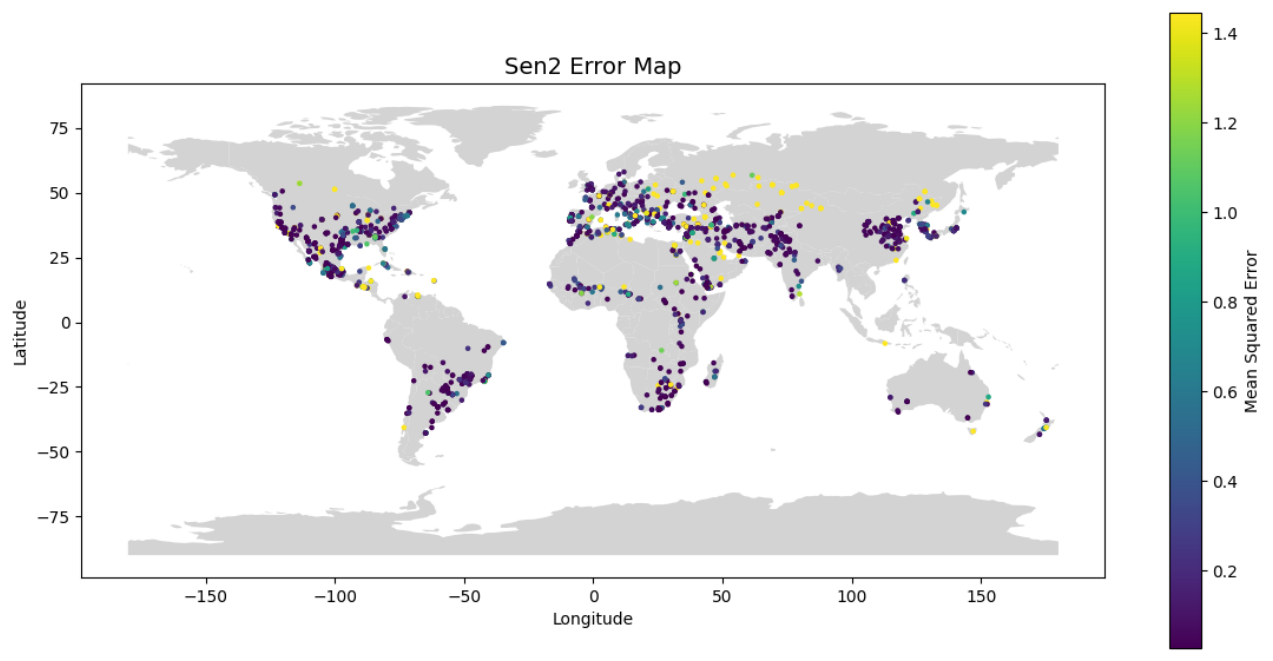}
    \caption{Spatial distribution of mean squared errors of the S-2 tokenizer on the validation set of the pretraining data.}
    \label{fig:s2-error-map}
\end{figure}

\begin{figure}  
    \centering 
    \includegraphics[width=\linewidth]{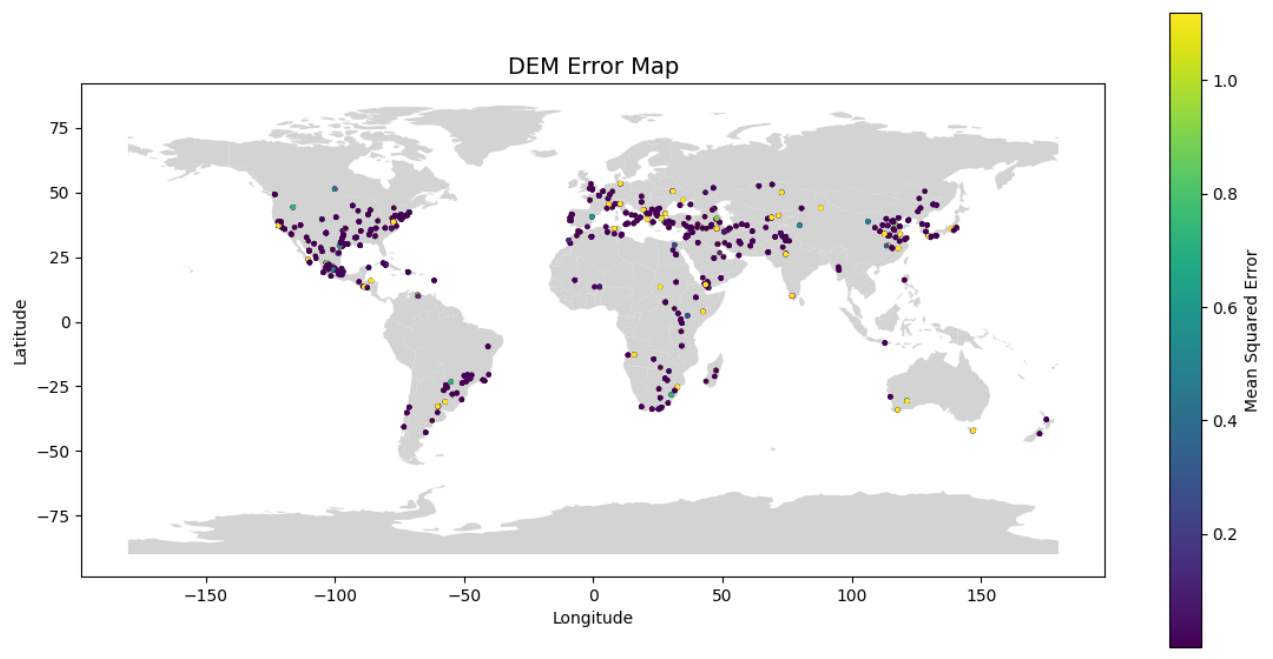}
    \caption{Spatial distribution of mean squared errors of the DEM tokenizer on the validation set of the pretraining data.}
    \label{fig:dem-error-map} 
\end{figure}

In Figures~\ref{fig:s1-error-map} to \ref{fig:dem-error-map}, we provide an overview on the spatial distributions of the S-1 GRD, S-2 L2A, and DEM tokenizer on the validation data of the SSL4EO-S12 subset which is focused on urban areas and therefore relevant for many downstream applications. Overall, we observe low MSE errors and particularly low deviation across geographic regions. For optical S-2 data, we observe minor difficulties in reconstructing images from Northern Asia, which we manually investigated. Overall, the vast majority of those samples are depicting snowy/icy conditions that have very high reflectance values of up to 12,000 compared to a normal range of [0, 255] in RGB data. On those long tail distribution samples, the S-2 tokenizer naturally has more difficulties. 

\textbf{S1-tokenizer quantitative analyses.}    
In the following, we pay particular attention to the performance of the radar S-1 tokenizer, which might be more challenging to train on a reconstruction task due to the inherent speckle noise in radar satellite data.    
We therefore evaluate the reconstructions of the S-1 tokenizer using the structural similarity index (SSIM) and peak signal-to-noise ratio (PSNR).  
Both input and reconstruction for S-1 are in a dB scale. 
In addition to S-1 evaluation metrics being computed in the dB space in Table~\ref{tab:sarevalresults}, they also are calculated in the denormalized space. On the contrary, the S-2 evaluation metrics are computed in the normalized space.

We give a more extensive background on radar data in the following for interested readers and non-EO experts. Reconstructing realistic and accurate synthetic aperture radar (SAR) S-1 VV and VH data is challenging due to factors inherent in the specific characteristics of SAR and the S-1 mission. SAR data is affected by complex interactions between the radar signal and Earth's surface. SAR is based on radar backscatter, which is influenced by surface roughness and moisture content. The interaction of radar waves with different surfaces, including vegetation structure and urban environments, can produce complex backscatter patterns.   
The two polarizations, VV and VH, capture different scattering mechanisms: VV is sensitive to surface roughness and vegetation, while VH captures cross-polarized interactions that are influenced by surface and volumetric features \cite{ChenweiWang, PrabhishekSingh, KoheiArai}.            
In addition, SAR inherently contains speckle noise, which obscures fine details, making it difficult to extract accurate information.     
To evaluate the SAR data tokenizers of TerraMind, we employ various evaluation metrics to assess quality and accuracy. We compute the MAE and RMSE for quantifying pixel-level differences, the SSIM to compare image structural content, and the PSNR \cite{ZWang, XinyuBai, AHore}.      

\color{black}
Table~\ref{tab:sarevalresults} presents the quantitative evaluation of the TerraMind tokenizer reconstructions across multiple modalities. The results show a reasonable reconstruction performance for optical data, indicating both structural and perceptual fidelity. For radar modalities, S-1 GRD and S-1 RTC achieve comparable PSNR values, though SSIM scores are lower, suggesting that while the reconstructions are visually plausible, they exhibit moderate structural deviations. In addition to these quantitative metrics, we also conducted qualitative assessments through visual inspection to identify artifacts and inconsistencies not captured by numerical scores alone.


\begin{table}[h]                   
    \centering     
    \small 
    \begin{tabular}{lcccc}
    \toprule
    \textbf{Modality} & \textbf{MAE} & \textbf{RMSE} & \textbf{SSIM} & \textbf{PSNR} \\
    \midrule
    S-1 GRD & 2.403 & 3.220 & 0.565 & 30.291 \\
    S-1 RTC & 2.216 & 2.888 & 0.466 & 30.389 \\
    S-2 L2A & 0.055 & 0.134 & 0.851 & 27.439 \\
    DEM   & 170.7 & 737.2 & 0.974 & 20.712 \\
    NDVI  & 0.091 & 0.168 & 0.647 & 21.517 \\
    \bottomrule
    \end{tabular}
    \caption{Evaluation of SAR VV and VH and S-2 reconstructions by the TerraMind tokenizers using MSE $\downarrow$, SSIM $\uparrow$ and PSNR $\uparrow$ on the validation dataset of the SSL4EO-S12 subset (8.5k samples).}    
    \label{tab:sarevalresults}      
\end{table}
\normalcolor

\section{Additional experiments} 

In the following, we provide additional experiments, especially with regard to the quality of the latent space and the full finetuning performance. To understand the quality of the latent space, we compute performances of nearest neighbor approaches for image classification tasks or using prototypical neural networks. We assess the performance of full finetuning by comparing with end-to-end trained, task-specific models like U-Nets and ViTs. We additionally compare the quality of the generations with the pseudo-labels used to pretrain TerraMind in an ablation experiment in a zero-shot setup.

\subsection{Geolocation prediction}

To better understand how TerraMind assigns geolocations, we further employ a Monte-Carlo sampling on the latitude-longitude grid for an optical tile from the validation data in Figure~\ref{fig:monte-carlo-val}. We observe that while TerraMind is not predicting the correct geolocation (\textcolor{red}{$\bullet$}), there is a very high likelihood that the predicted geolocation is one of the adjacent grid points that have been seen during pretraining (\textcolor{black}{$\bullet$}). \color{black}This result suggests that even for data from unseen geolocations, TerraMind remembers similar samples from the pretraining data (\textcolor{green}{$\bullet$}) and returns the geolocation of the samples with high similarity. This capability paired with the global pretraining of TerraMind suggests that geo-localization of data from unseen locations is possible but determined by the similarity to images from adjacent locations.\normalcolor

\definecolor{darkgreen}{rgb}{0.0, 0.69, 0.0}

\begin{figure}[tbh]
        \centering
        \includegraphics[width=0.8\linewidth]{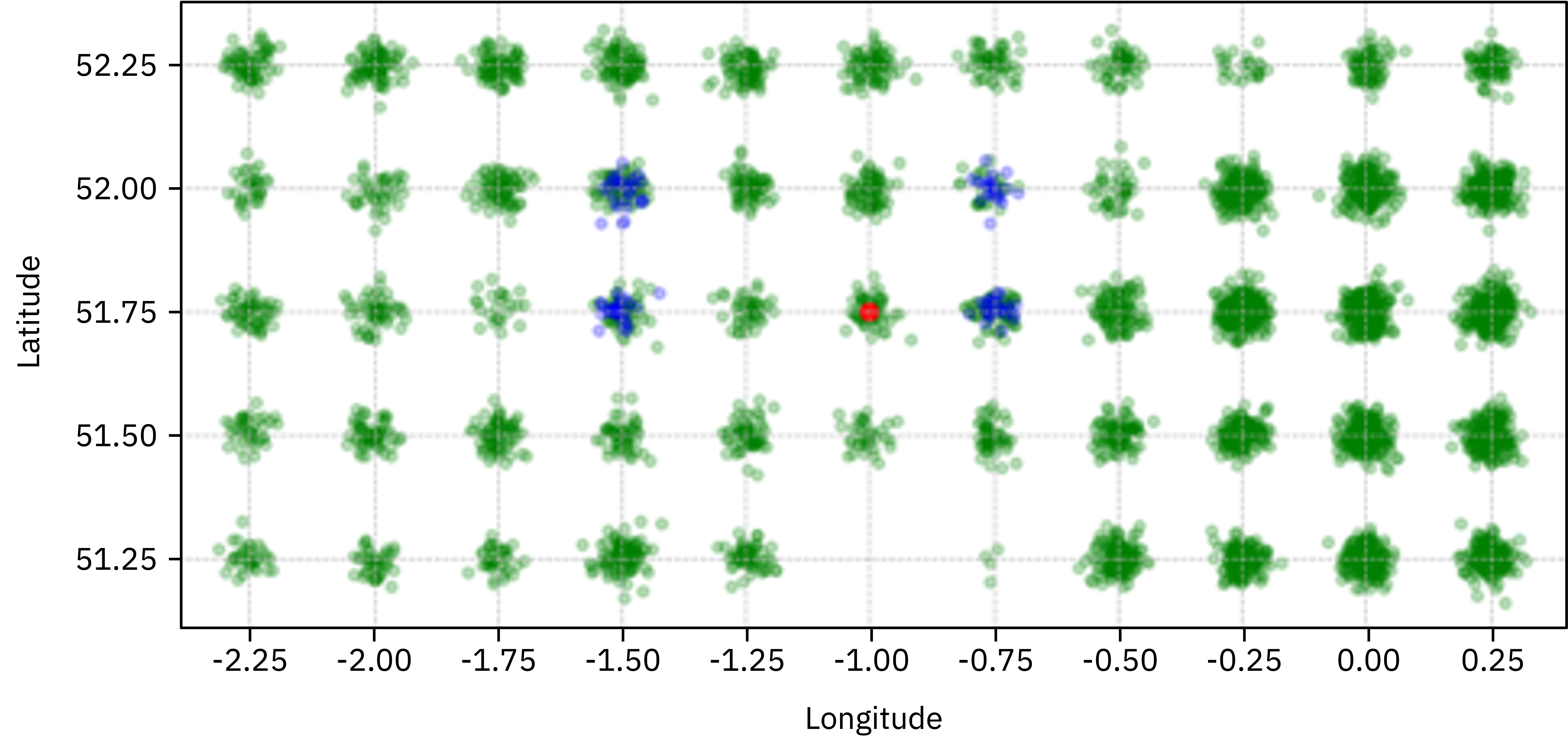}
        \caption{\textcolor{black}{Distribution of predicted geo-locations for an optical S-2 L2A sample from the validation set. \textcolor{red}{$\bullet$} is the correct location, \textcolor{black}{$\bullet$} are Monte-Carlo sampled locations from TerraMind, \textcolor{darkgreen}{$\bullet$} represents the distribution of training locations. TerraMind's geo-localization seems to be based on similar optical samples in the training dataset for which TerraMind then outputs the geolocation.}}
        \label{fig:monte-carlo-val}
\end{figure}

\color{black}
We further extend the analysis of Figure~\ref{fig:lulc_geo_loc} by additionally prompting the model for likely locations of urban areas. Overall, we observe that the model correctly identifies many densly populated areas across the globe. We also note overpredictions in, for example, North Africa and middle-east. This observation suggests that the model might confuse bare land and urban areas in these regions. 
\normalcolor

\begin{figure}[h]
    \centering
        \centering
        \includegraphics[width=0.8\linewidth]{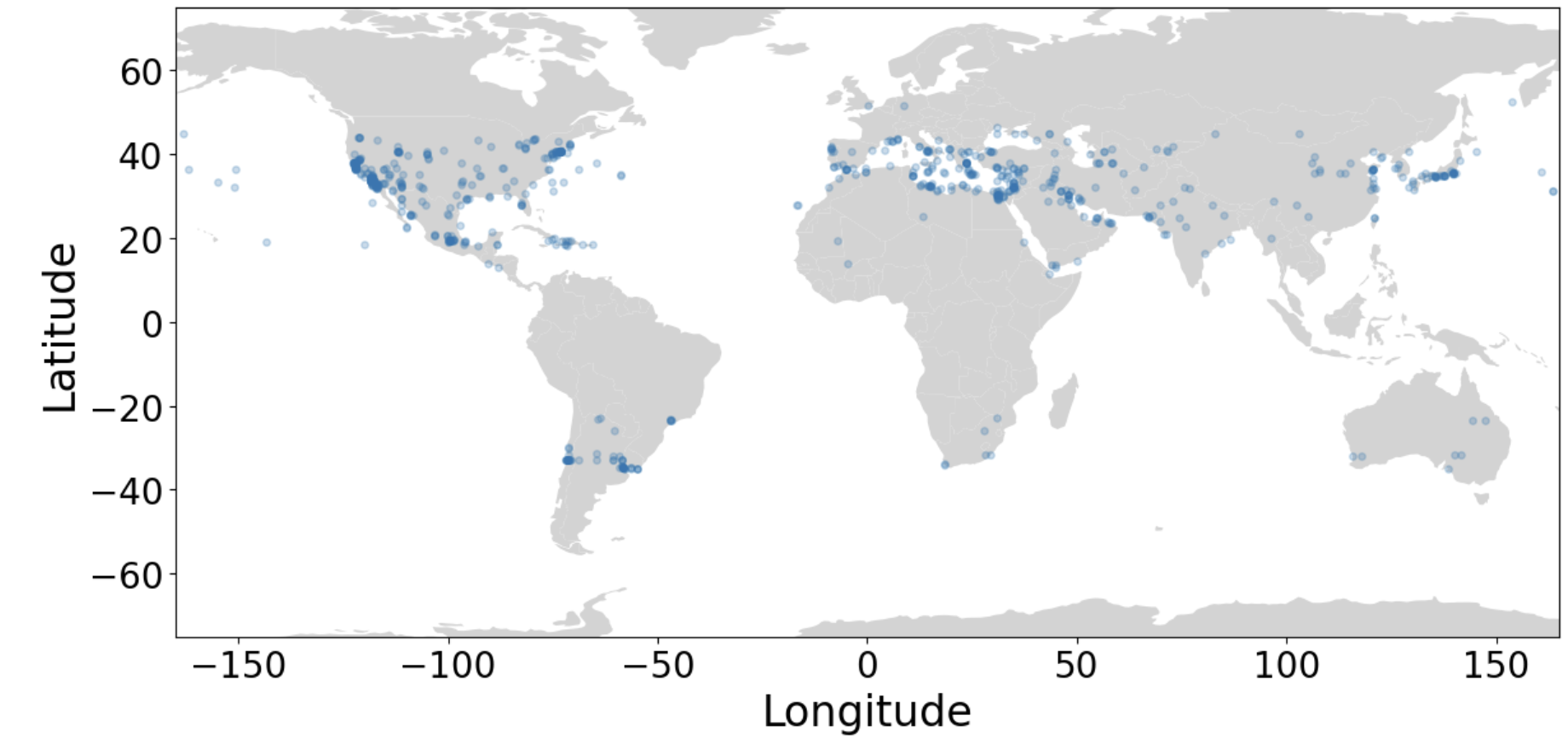}
        \caption{\textcolor{black}{Prediction distribution of the land use class ``urban'' with a sampling temperature of $T=1.0$. TerraMind has a reasonable internal representation of the geolocation of specific contexts, like land use classes.}}
        \label{fig:urban_geo_loc}
\end{figure}

\subsection{Few-shot experiments}  

We present additional few-shot experiments with the EuroSAT and METER-ML dataset in Table~\ref{tab:few_shot_all}. We use the embeddings of the pre-trained encoders without any additional fine-tuning. The patch embeddings of each image are averaged for image-level classification tasks.

The experiments include four different few-shot settings with varying numbers of examples and classes. 5-way refers to sampling five classes per run, while full-way describes experiments with all dataset classes per run. 1-shot and 5-shot indicate that one or five images are sampled for each class per run. 5-shot experiments with five support samples per class are using Prototypical Networks~\cite{snell2017prototypical} for classification. This approach averages the embeddings of the selected labeled images (support set) and classifies the target images (query set) based on the class prototype with the lowest Euclidean distance from each sample. In the 1-shot setting, Prototypical Networks are mathematically equal to 1-Nearest-Neighbor classification. We refer to the original paper for details~\cite{snell2017prototypical}. Different from literature, we evaluate each run on the full test set instead of subsampling query images.

TerraMind performs best on both datasets, outperforming all other geospatial foundation models as well as the CLIP vision encoder~\cite{clip}. Interestingly, the base version leads to overall better results than the large model. Similarly, Prithvi's smaller 1.0 version has comparable results to its larger 2.0 300M version, indicating that model size has only a limited effect on few-shot performance. 

In addition to S-2 L1C, the METER-ML dataset provides high resolution RGB images from NAIP with 1\,m resolution. Only CLIP and TerraMind can process RGB images without any fine-tuning. While CLIP profits largely from the higher resolution inputs, TerraMind only performs marginally better and sometimes worse than with multispectral S-2 data. Notice that TerraMind shows similar performance gaps as CLIP when comparing NAIP data to S-2 RGB. This indicates that additional multispectral channels have a comparable effect on few-shot performance as high-resolution images.

\begin{table*}[tbh]
    \centering
    \small
    \setlength{\tabcolsep}{2pt}
    \begin{tabularx}{\textwidth}{lcCCCC | CCCC }
    \toprule
    & & \multicolumn{4}{c}{EuroSAT} & \multicolumn{4}{c}{METER-ML} \\
    Model & Input & 5-way 1-shot & 5-way 5-shot & full-way 1-shot & full-way 5-shot & 5-way 1-shot & 5-way 5-shot & full-way 1-shot & full-way 5-shot \\
    \midrule
    {CLIP-ViT-B/16} & {S-2 RGB} &  57.00 & 70.72 & 43.92 & 58.30 & 29.15 & 37.44 & 23.13 & 30.53 \\
    {CLIP-ViT-B/16} & {NAIP} & – & – & – & – & 32.01 & 42.35 & 25.66 & 35.81 \\
    {DeCUR} & {S-2 L1C} & 50.54 & 64.35 & 37.53 & 50.82 & 27.87 & 33.64 & 20.95 & 27.21\\
    {Prithvi 1.0 100M} & {S-2 L1C} & 60.11 & 73.29 & 46.86 & 60.66 & 26.08 & 35.81 & 22.33 & 29.21 \\
    {Prithvi 2.0 300M} & {S-2 L1C}  & 61.06 & 73.21 & 47.47 & 60.47 & 28.26 & 36.13 & 22.52 & 29.59 \\
    \midrule
    {{TerraMindv1-B}} & {S-2 L1C} & \textbf{70.83} & \textbf{87.94} & \textbf{57.48} & \textbf{79.66} & \textbf{33.90} & 43.89 & \textbf{26.85} & \underline{37.41} \\
    {{TerraMindv1-B}} & {NAIP} & – & – & – & – & 32.23 & \underline{44.75} & 25.53 & 37.85 \\
    {{TerraMindv1-L}} & {S-2 L1C} & \underline{70.07} & \underline{86.29} & \underline{56.58} & \underline{77.39} & \underline{33.09} & 42.72 & \underline{26.02} & 36.34 \\
    {{TerraMindv1-L}} & {NAIP} & – & – & – & – & 32.59 & \textbf{44.99} & 25.94 & \textbf{38.29} \\
    \bottomrule
    \end{tabularx}
    \caption{Few-shot classification results on EuroSAT and METER-ML measured in mean accuracy~$\uparrow$ averaged over 200 runs. 5-way refers to five randomly sampled classes per run, which is a default setting used in few-shot learning. Full-way refers to sampling all dataset classes, i.e., ten EuroSAT classes and seven METER-ML classes. We highlight the best two models in bold and underlined.}
    \label{tab:few_shot_all}
\end{table*}

\subsection{Finetuning comparisons with baseline models}

Since the first approaches to foundation models for Earth observations, experts in the field discuss on the usability of such models compared to task-specific models that are trained for each application individually. Recent benchmark results suggested that task-specific models, like U-Nets, often outperform finetuned GFMs \cite{marsocci2024pangaea}. We therefore additionally investigate how TerraMind compares with task-specific U-Nets and ViT models following the PANGAEA evaluation protocol in Table~\ref{tab:pangaea-full-results}. As adviced by the authors of PANGAEA, we again report results on nine of the eleven datasets as we could not reproduce the performance on the remaining two datasets. The task-specific models are trained from scratch for each individual task, while all GFMs including TerraMind are finetuned with a frozen encoder and an UperNet head. 
Overall, our results demonstrate that TerraMindv1-B outperforms task-specific UNet and ViT models across the PANGAEA benchmark in both unimodal and multimodal settings by 1pp avg. mIoU and 4pp avg. mIoU respectively. In multimodal settings, the improvement peaks to 4.5pp improvement of TerraMindv1-B over task-specific U-Nets. 
To the best of our knowledge, this is the first time a GFM model outperforms task-specific models on a global benchmark. 

In addition, we observe that for most datasets, TerraMindv1-B outperforms TerraMindv1-B-single. This demonstrates the benefit from scaling in the data and feature dimension--i.e., leveraging dual-scale feature representations on a pixel level and a token level.

\subsection{Comparing generations and pseudo-labels}    

We evaluate the model generations for modalities where we used pseudo-labels as input data. For example, in initial experiments with TerraMindv1-B-single, we leverage Google's DynamicWorld model to pseudo-label LULC maps which we use as input to the model.  
In the following experiment in Table~\ref{tab:sen1floods11_zeroshot_results}, we test the performance of the DynamicWorld model against the generations of TerraMind. 
Overall, we observe that while finetuned TerraMindv1-B-single outperforms DynamicWorld, the generation of TerraMind does not surpass the inference results of DynamicWorld.

\begin{table}[tbh]      
    \centering  
    \small
    \begin{tabular}{lcc}
        \toprule
        \textbf{Approach} & \textbf{Input} & \textbf{IoU$_{Water}$} \\
        \midrule
        TerraMindv1-B-single & S-2 L1C & 69.87 \\
        Dynamic World pseudo-labeling & S-2 L1C & 71.98 \\
        \midrule
        TerraMindv1-B-single finetuning & S-2 L1C & \textbf{76.32} \\
        \bottomrule
    \end{tabular}
    \caption{Results on the Sen1Floods11 test set comparing flood maps derived from TerraMind's out-of-the-box LULC generations to those derived from LULC pseudo-labeling with Dynamic World. The results are inferior to those obtained by fine-tuning a specialized model for this downstream task, which is expected.}
    \label{tab:sen1floods11_zeroshot_results} 
\end{table}

\subsection{TiM tuning for crop mapping}

We further investigate the relevance of TiM tuning for crop type mapping in order to understand the relevance of generating artificial data for more finegrained segmentation tasks. That means, we generate artificial LULC data which includes agricultural crop as a \textit{single class} and investigate whether this additional information helps to segment nine different types of crops in satellite images. We experiment with the South Africa Crop Type Mapping dataset (\url{https://source.coop/esa/fusion-competition}) and present the results in Table~\ref{tab:sacrop}. Overall, we observe that TiM tuning improves the performance by around 1pp. That means that even though the generated artificial data does not include further information on the location and shape of certain crops, the information on where to expect crop land in general helps to guide the model to an improved performance.

\begin{table}[h] 
    \centering
    \small
    \begin{tabularx}{\linewidth}{lcC}
        \toprule
         & {Input} & mIoU \\
        \midrule
        TerraMindv1-B  & S-2 & 41.87 \\
        \textbf{TerraMindv1-B TiM}  & S-2 \textit{+ gen. LULC} &  \textbf{42.74} \\
        \bottomrule
    \end{tabularx}
    \caption{Thinking-in-modalities (TiM) tuning compared with standard full fine-tuning approaches on the SA Crop dataset.}
    \label{tab:sacrop}
\end{table}

\section{Any-to-any generation}  

\begin{figure*}[htb]
    \centering
    \includegraphics[width=\linewidth]{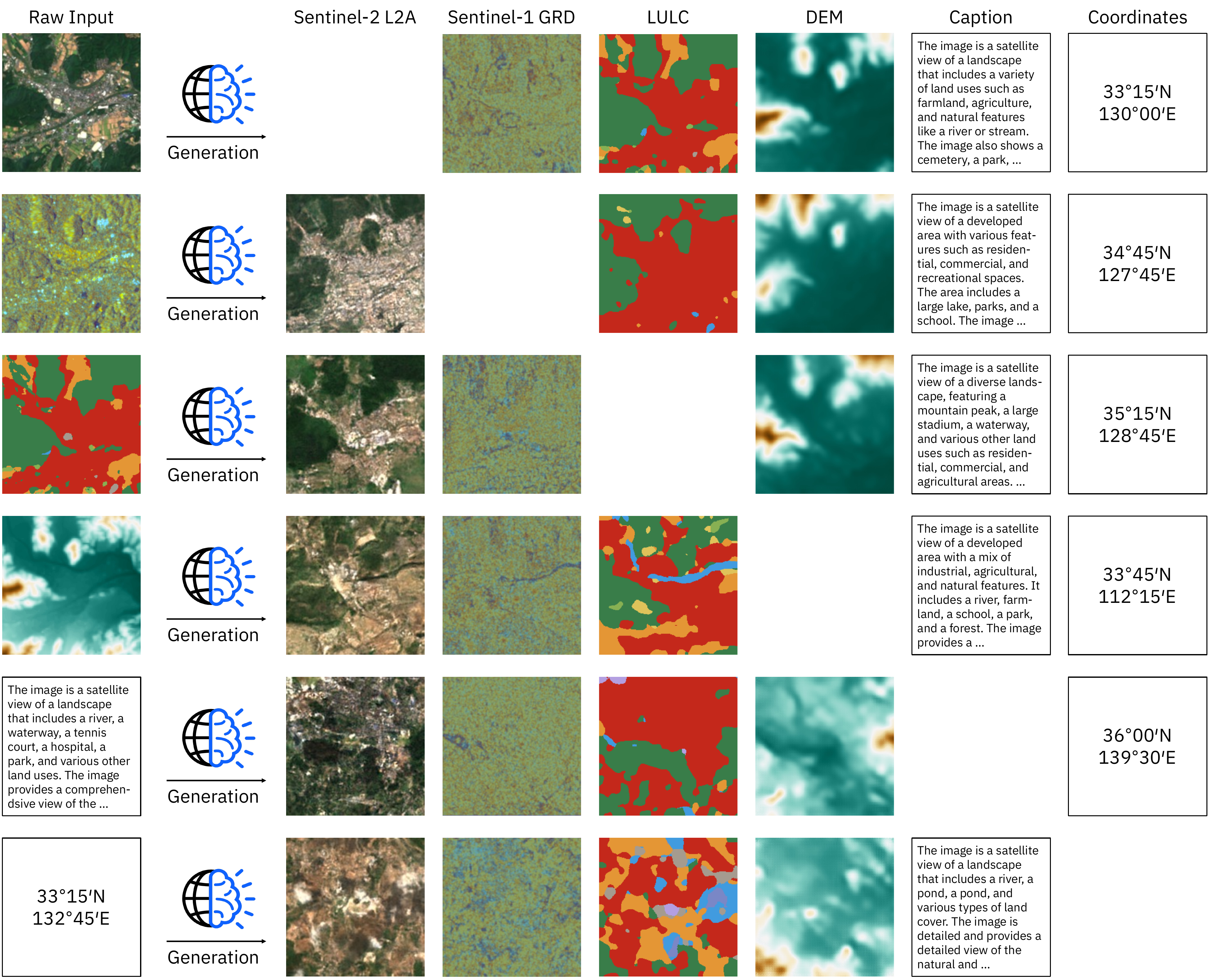}
    \caption{Any-to-any generation example of TerraMindv1-B-single. Fine-grained input like optical and radar achieve particularly good performances.}
    \label{fig:any-to-any-example}
\end{figure*}

In Figure~\ref{fig:any-to-any-example}, we provide an example of any-to-any generation on four image-like modalities and two sequence-like modalities. Overall, we observe that when we start from modalities with high information content (e.g., fine-grained image-like modalities), the reconstructions are particularly good. Even with less information content, the model is able to generate consistent artificial data. However, we can clearly observe that the quality compared to the ground truth (represented by the input in the left of the figure) is decreasing. Finally, it is interesting to see how artefacts are introduced by the model when starting from lower information content in the input. For example, when promting TerraMind to generate data from DEM input, we observe that the model pays significant attention to the darker streams in the DEM image, which are later generated as a river in LULC.

\begin{figure}[h]       
    \centering      
    \includegraphics[width=1\linewidth]{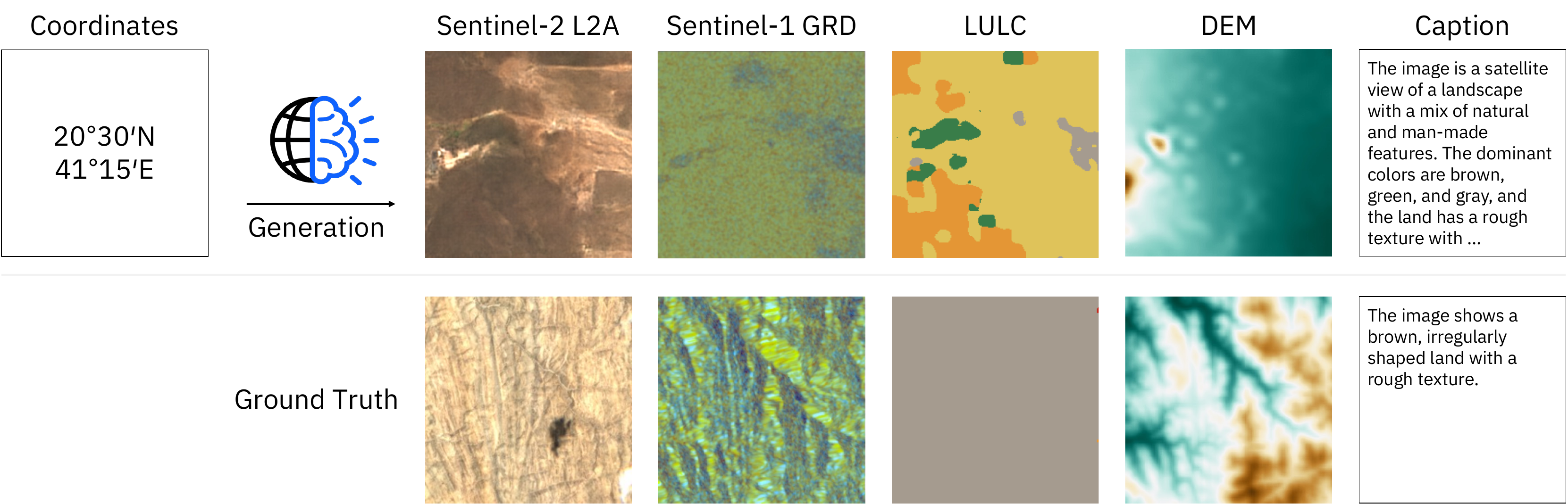}
    \caption{\textcolor{black}{Randomly selected chained generation example with uni-modal geo-location input data. Top row is artificially generated data by TerraMind, buttom row represents a ground truth sample at this grid location, respectively.}}
    \label{fig:desert-generation}
\end{figure}

\color{black}
While we expect to see accurate generations from information-rich modalities like optical data, it is particularly interesting to understand how TerraMind deals with low information content. Therefore, we prompt TerraMind to generate a subset of modalities starting from the geolocation in Figure~\ref{fig:desert-generation}. Interestingly, for a geolocation from the middle-east, the model generates an optical image that resembles a desert. While the generated optical image is based on the right context, the actual structure is unsurprisingly different from the ground truth. Based on the chained generation, this difference ripples down across all other modalities as well causing consistent but inaccurate generations. This example emphasizes the relevance of access to information-rich, fine-grained features to facilitate accurate generations. 

\color{black}Next to the evaluation of raw, pixel-level input in Table~\ref{tab:generation_performance}, we further evaluate the generation quality using tokenized input in Table~\ref{tab:tokenized-performance}. Interestingly, we observe only minor reduction in performance compared to pixel-level input even though the tokenized representations are compressed significantly (up to 3000x for S-2 L2A). Overall, our results suggest that leveraging tokenized inputs can be a reasonable alternative to leveraging pixel-level data for the generation of artificial data with TerraMind.

\begin{table*}[h]
\centering
\color{black}
\arrayrulecolor{black}
\begin{tabular}{lcccc}
\toprule
\textbf{Modalities} & \textbf{MAE} & \textbf{RMSE} & \textbf{SSIM} & \textbf{PSNR} \\
\midrule
Tokenized S-2 L2A $\rightarrow$ S-1 GRD & 3.3180 & 4.3309 & 0.5131 & 27.715 \\
Tokenized S-2 L2A $\rightarrow$ S-1 RTC & 3.0544 & 3.9178 & 0.4131 & 27.739 \\
Tokenized S-2 L2A $\rightarrow$ DEM   & 572.5 & 1040.6 & 0.5728 & 17.718 \\\midrule
Tokenized S-1 GRD $\rightarrow$ S-2 L2A & 0.0820 & 0.1238 & 0.7182 & 25.630 \\
Tokenized S-1 GRD $\rightarrow$ NDVI  & 0.1949 & 0.2425 & 0.4124 & 18.324 \\
Tokenized S-1 GRD $\rightarrow$ DEM   & 327.4 & 550.3 & 0.7271 & 16.008 \\\midrule
Tokenized S-1 RTC $\rightarrow$ S-2 L2A & 0.1195 & 0.1935 & 0.6638 & 24.266 \\
Tokenized S-1 RTC $\rightarrow$ NDVI  & 0.1895 & 0.2348 & 0.4500 & 18.606 \\
Tokenized S-1 RTC $\rightarrow$ DEM   & 457.9 & 851.6 & 0.7095 & 19.457 \\
\bottomrule
\end{tabular}
\caption{\color{black}Performance of TerraMind on tokenized inputs using 10 diffusion steps. Metrics include MAE $\downarrow$, RMSE $\downarrow$, PSNR $\uparrow$, and SSIM $\uparrow$.\normalcolor}
\label{tab:tokenized-performance}
\end{table*}
\arrayrulecolor{black}
\normalcolor

\subsection{\textcolor{black}{Large-scale generations}}

\color{black}In Figures~\ref{fig:singapore-large_tile} and \ref{fig:santiago-large_tile}, we provide additional qualitative results for large-tile generations at the example of Singapore. Specifically, we leverage a 35.5km $\times$ 69.5km optical S-2 L2A tile as input and iteratively generate overlapping 224x224 pixel generations for S-1 RTC, S-1 GRD, NDVI, and LULC. In the overlapping areas, we apply the mean of all generations in order to enhance the spatial conciseness of the generations.  
TerraMind consistently removes the clouds in the S-1 generations. It makes assumptions for hidden areas, which are look accurate for large features like water bodies or the shore line. Other features like airports or ships are also clearly visible in the S-1 and NDVI generations.







    
    

\begin{figure*}[p]
    \centering
    \begin{subfigure}[b]{\textwidth}
        \includegraphics[width=\textwidth]{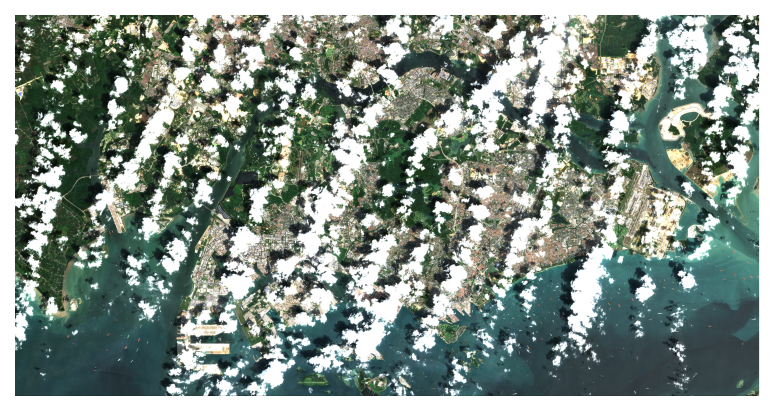}
        \caption{\textcolor{black}{Input: S-2 L2A data from Singapore captured January 9th, 2025.}}
    \end{subfigure}
    \begin{subfigure}[b]{\textwidth}
        \includegraphics[width=\textwidth]{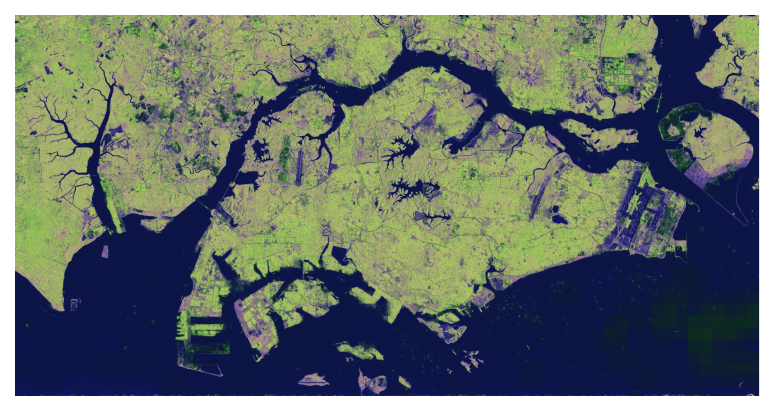}
        \caption{\textcolor{black}{Generation: TerraMind output for S-1 composition}}
    \end{subfigure}
    \caption{\textcolor{black}{Large-tile generations of TerraMind for Singapore (1/1)}}
    \label{fig:singapore-large_tile}
\end{figure*}

\begin{figure*}[p]
    \ContinuedFloat
    \centering
    \begin{subfigure}[b]{\textwidth}
        \includegraphics[width=\textwidth]{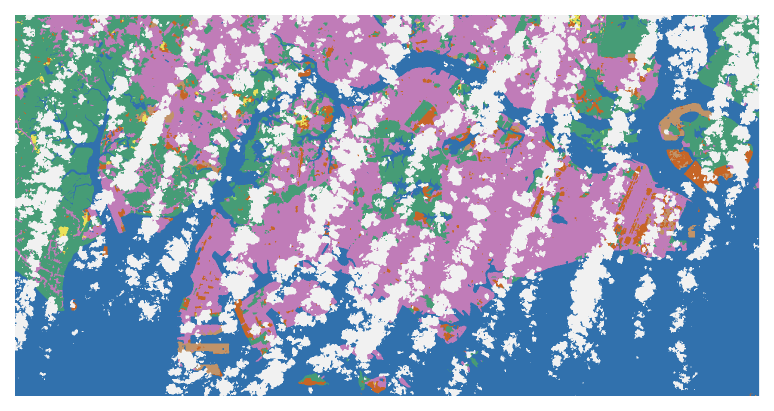}
        \caption{\textcolor{black}{Generation: TerraMind output for LULC}}
    \end{subfigure}
    \caption{\textcolor{black}{Large-tile generations of TerraMind for Singapore (2/2)}}
\end{figure*}

\begin{figure*}[p]
    \centering
    \begin{subfigure}[b]{\textwidth}
        \includegraphics[width=\textwidth]{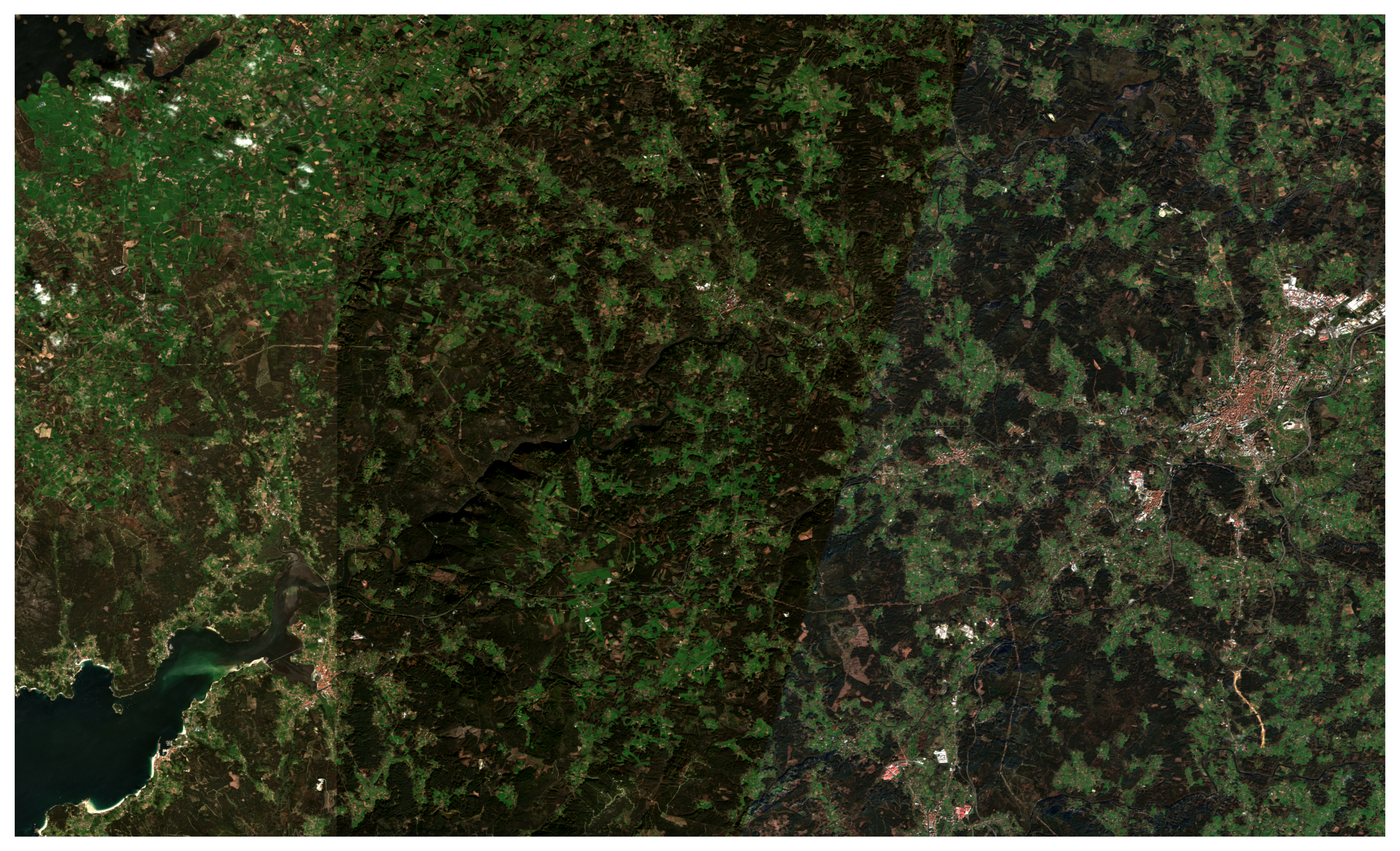}
        \caption{\textcolor{black}{Input: S-2 L2A data from Santiago de Compostela.}}
    \end{subfigure}
    \begin{subfigure}[b]{\textwidth}
        \includegraphics[width=\textwidth]{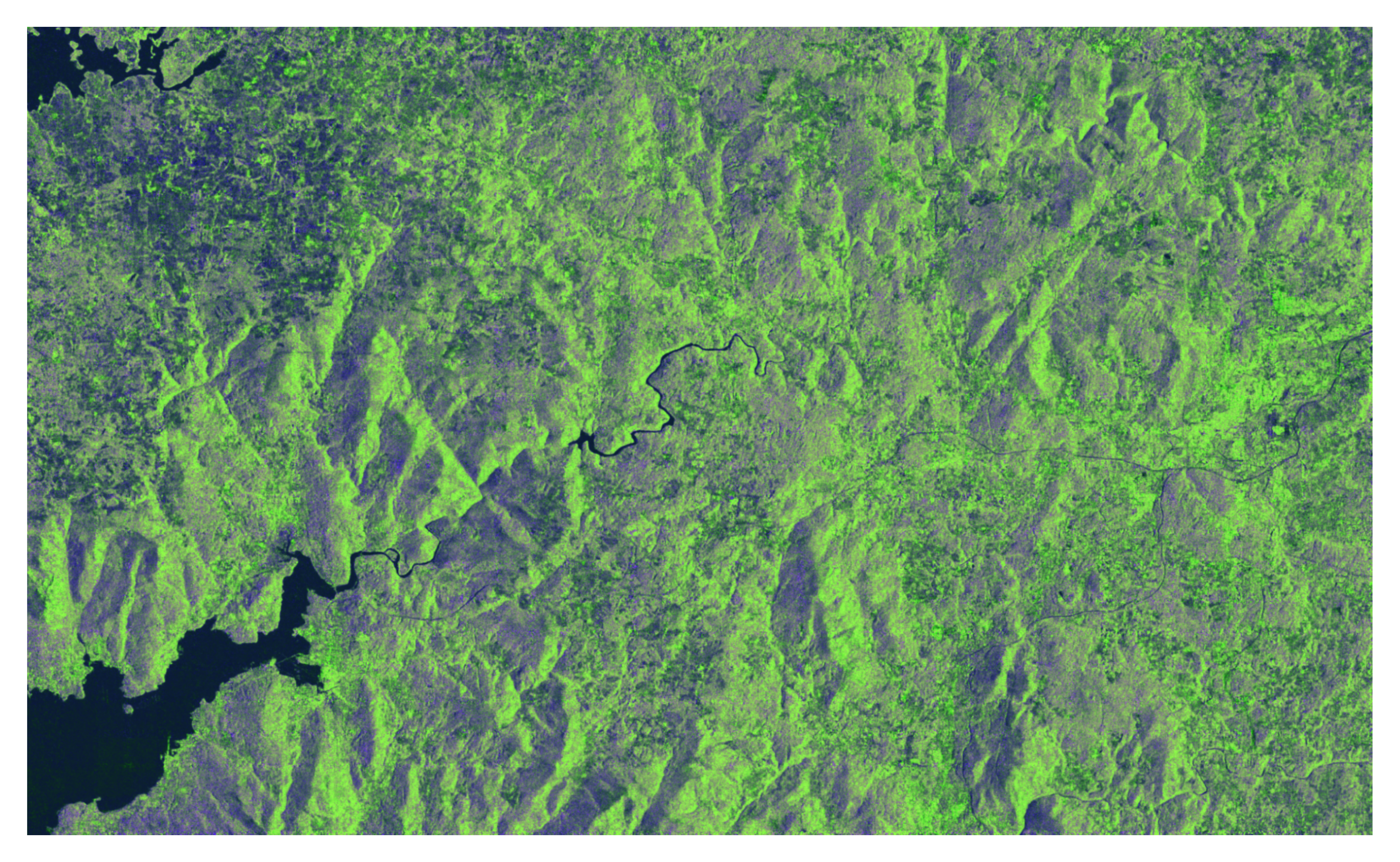}
        \caption{\textcolor{black}{Generation: TerraMind output for S-1 GRD composition}}
    \end{subfigure}
    \caption{\textcolor{black}{Large-tile generations of TerraMind for Santiago de Compostela (1/3)}}
    \label{fig:santiago-large_tile}
\end{figure*}

\begin{figure*}[p]
    \ContinuedFloat
    \centering
    \begin{subfigure}[b]{\textwidth}
        \includegraphics[width=\textwidth]{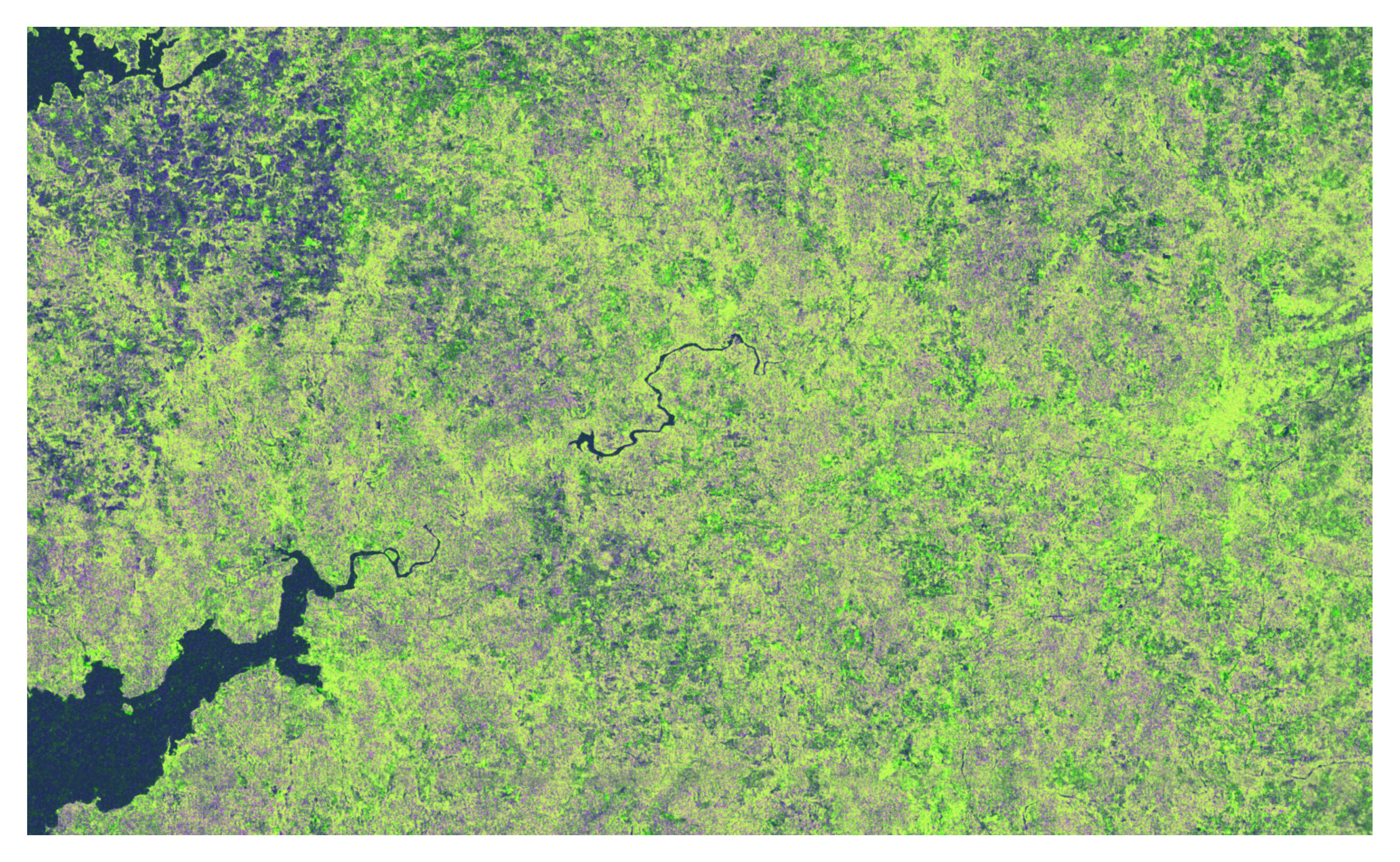}
        \caption{\textcolor{black}{TerraMind generation for S-1 RTC composition}}
    \end{subfigure}
    \begin{subfigure}[b]{\textwidth}
        \includegraphics[width=\textwidth]{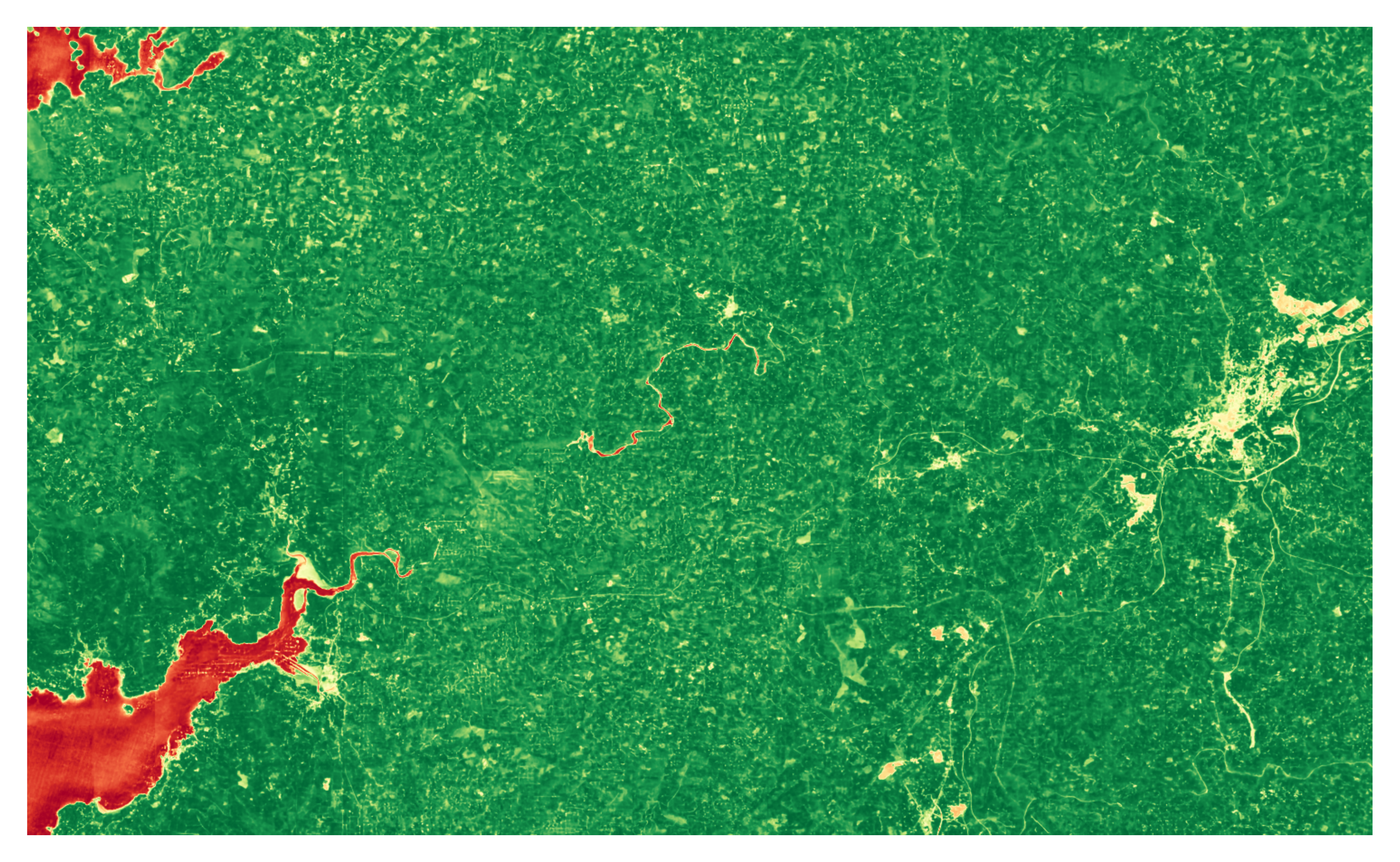}
        \caption{\textcolor{black}{Generation: TerraMind output for vegetation}}
    \end{subfigure}
    \caption{\textcolor{black}{Large-tile generations of TerraMind for Santiago de Compostela (2/3)}}
\end{figure*}

\begin{figure*}[p]
    \ContinuedFloat
    \centering
    \begin{subfigure}[b]{\textwidth}
            \includegraphics[width=\textwidth]{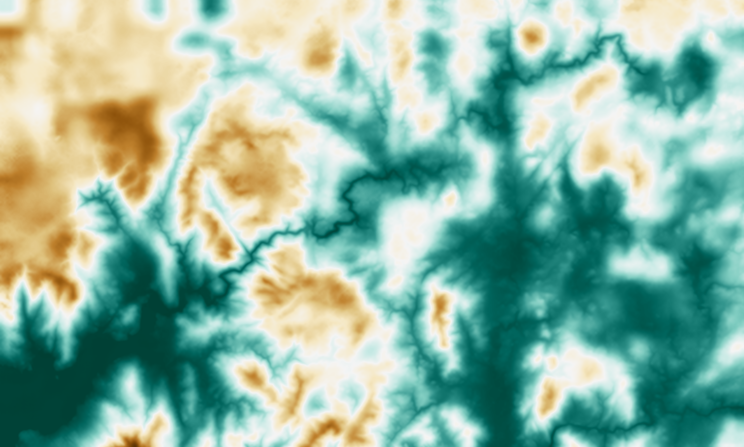}
            \caption{\textcolor{black}{Generation: TerraMind output for digital elevation}}
    \end{subfigure}
    \caption{\textcolor{black}{Large-tile generations of TerraMind for Santiago de Compostela (3/3)}}
\end{figure*}
\appendix

\end{document}